\documentclass{article}

\usepackage{microtype}
\usepackage{graphicx}
\usepackage{booktabs} 

\usepackage{hyperref}

\usepackage[accepted]{icml2024}

\usepackage{amsmath}
\usepackage{amssymb}
\usepackage{mathtools}
\usepackage{amsthm}

\usepackage[capitalize,noabbrev]{cleveref}

\theoremstyle{plain}

\theoremstyle{definition}

\theoremstyle{remark}

\usepackage[textsize=tiny]{todonotes}

\usepackage{svg}
\usepackage{subcaption}
\usepackage{multirow}
\usepackage{dsfont}
\newcommand{\topk}{top-$k$ }
\newcommand{\Topk}{Top-$k$ }

\icmltitlerunning{Conformal Prediction Sets Improve Human Decision Making}

\begin{document}

\twocolumn[
\icmltitle{Conformal Prediction Sets Improve Human Decision Making}


\begin{icmlauthorlist}
\icmlauthor{Jesse C. Cresswell}{L6}
\icmlauthor{Yi Sui}{L6}
\icmlauthor{Bhargava Kumar}{TDS}
\icmlauthor{No\"el Vouitsis}{L6}
\end{icmlauthorlist}
\icmlaffiliation{L6}{Layer 6 AI}
\icmlaffiliation{TDS}{TD Securities}
\icmlcorrespondingauthor{Jesse C. Cresswell}{jesse@layer6.ai}

\icmlkeywords{Conformal Prediction, Uncertainty Quantification}

\vskip 0.3in
]



\printAffiliationsAndNotice{}  

\begin{abstract}
In response to everyday queries, humans explicitly signal uncertainty and offer alternative answers when they are unsure. Machine learning models that output calibrated prediction sets through conformal prediction mimic this human behaviour; larger sets signal greater uncertainty while providing alternatives. In this work, we study the usefulness of conformal prediction sets as an aid for human decision making by conducting a pre-registered randomized controlled trial with conformal prediction sets provided to human subjects. With statistical significance, we find that when humans are given conformal prediction sets their accuracy on tasks improves compared to fixed-size prediction sets with the same coverage guarantee. The results show that quantifying model uncertainty with conformal prediction is helpful for human-in-the-loop decision making and human-AI teams.
\end{abstract}

\section{Introduction}\label{sec:intro}

\begin{figure}[t]
    \centering
    \includegraphics[scale=0.28]{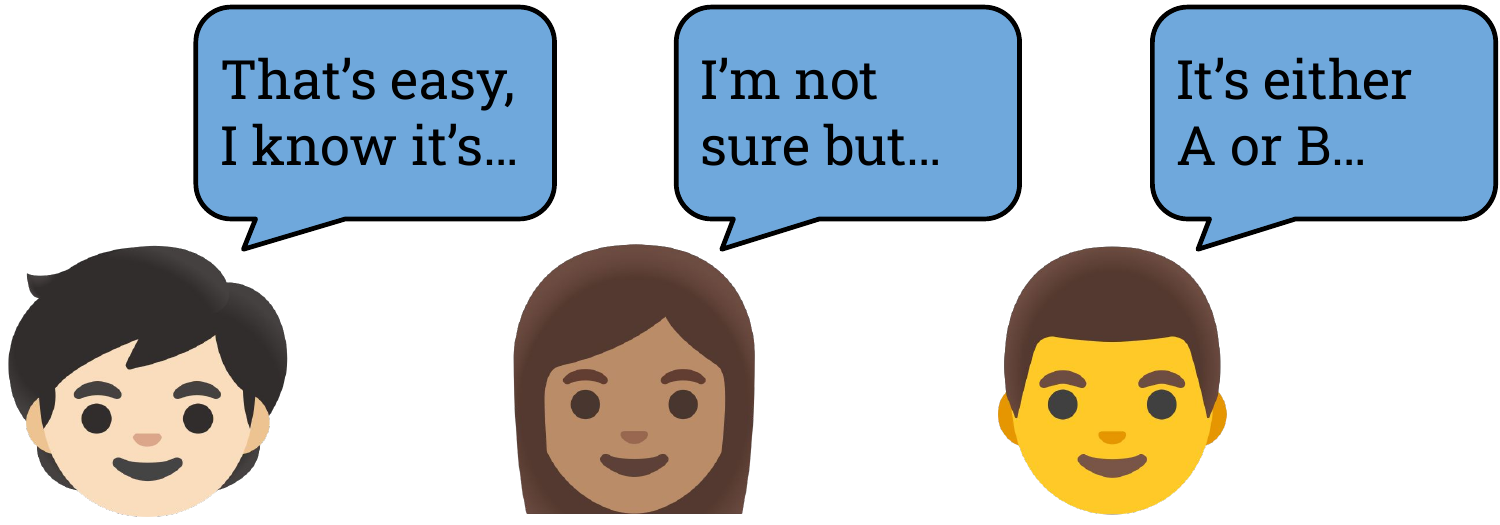}\\
    \vspace{13pt}
    \includegraphics[scale=0.42, trim={0, 0, 195, 4}, clip]{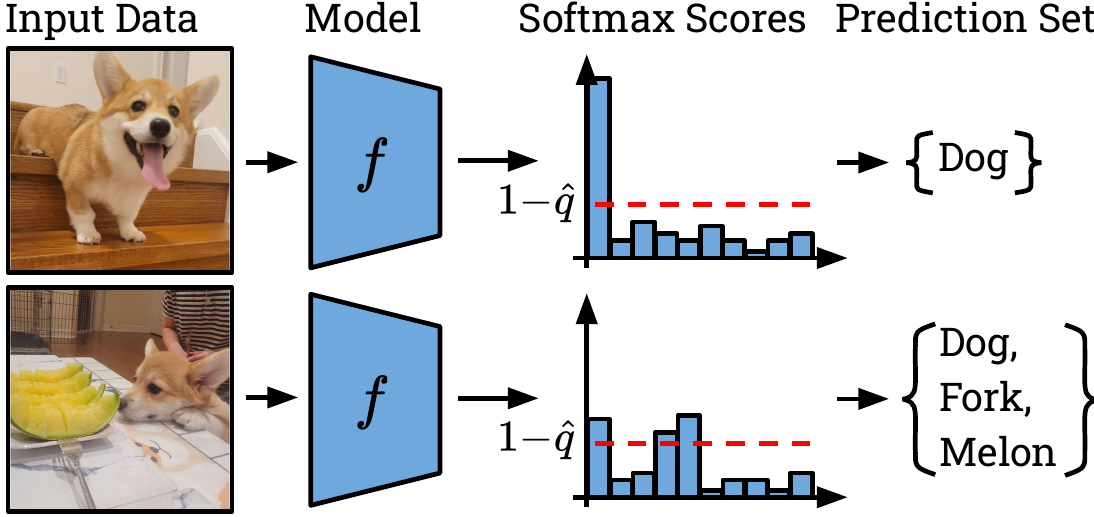}
    \caption{\textbf{Top:} Humans express uncertainty through explicit signalling, and by offering alternatives. \textbf{Bottom:} Conformal prediction allows machine learning models to do the same by outputting prediction sets with size calibrated to model uncertainty. Larger sets signal greater uncertainty and provide alternative answers.}
    \label{fig:human_uncertainty}
    \vspace{-15pt}
\end{figure}

When answering questions, humans naturally provide information on how confident we are in our answers. If unsure, we \emph{signal uncertainty} to others \cite{smith1993questions}, and \emph{offer alternatives} \cite{berlyne1962uncertainty} (\autoref{fig:human_uncertainty}). On the other hand, standard methods for discriminative machine learning often output only a single answer without quantifying the uncertainty in the prediction \cite{abdar2021review}. The lack of uncertainty quantification and lack of alternative predictions greatly limits the usefulness of machine learning models for real-world decision making.

Prediction sets are a useful tool for augmenting the output of a discriminative model. Rather than outputting a single prediction, a model can output a \emph{set} of predictions, which may contain zero, one, or multiple possible values. Generally, prediction sets are constructed to contain the values that are most likely to be correct according to the model using some heuristic notion of confidence. Perhaps the most straightforward way to construct prediction sets in the classification context is by including the \topk values according to softmax outputs for a fixed $k$ \cite{grycko1993classification}. \Topk prediction sets are useful for providing alternatives and can even furnish statistical guarantees \cite{chzhen2021set}; however, they do not quantify uncertainty.

Conformal prediction \cite{vovk2005algorithmic} is a general-purpose method for transforming heuristic notions of uncertainty into rigorous ones through the use of \emph{calibrated} prediction sets. Conformal sets usually are designed so that the set size indicates how confident the model is about a particular input, with larger sets signalling more uncertainty (\autoref{fig:human_uncertainty}). The hallmark of conformal prediction is its risk control; conformal sets will only fail to contain the ground truth with a pre-specified error rate.
Furthermore, conformal prediction is widely applicable since it can be used with any pre-trained model, and does not rely on distributional assumptions, nor on infinite-data limits \cite{angelopoulos2022gentle}. It has been successfully applied in real-world settings for  
drug discovery \cite{eklund2015application, alvarsson2021predicting}, medical diagnosis \cite{zhan2020electronic}
, time-series forecasting \cite{stankeviciute2021conformal, xu2021conformal, zaffran2022adaptive}, and language modelling \cite{kumar2023conformal, ren2023robots}.

The main drawback of prediction sets, even conformally calibrated ones, is that decision making in the real world often requires us to act on a \emph{single} option, which we are not guaranteed to obtain through conformal prediction. Given the qualitative similarity between human expressions of uncertainty and conformal prediction, it is natural to rely on humans as the mechanism for converting prediction sets into decisions. Prediction sets can be provided to a human who uses the information along with their own judgement to make a final decision, with the expectation that the combination outperforms a human alone. Despite its reasonableness, this expectation has not been scientifically tested.

In this work, we conduct a pre-registered\footnote{The pre-registration is viewable at \href{https://osf.io/fkdhv}{osf.io/fkdhv}.}
randomized controlled trial to directly measure human performance using three classification tasks. We evaluate the benefits of both alternative suggestions and uncertainty quantification to human decision makers by providing three levels of assistance: no assistance (control), \topk sets, and conformal sets. Compared to the control, \topk sets provide a fixed number of likely alternatives to the human, while conformal sets also quantify uncertainty. Our main finding is that humans do leverage the uncertainty quantification provided by conformal sets. In all three independent tasks we find with statistical significance ($p\text{-values}<0.05$) that conformal prediction sets improve human accuracy compared to \topk sets, although providing prediction sets does not always speed up decision making compared to the control.

\section{Background and Related Work}\label{sec:background}

\subsection{Uncertainty Quantification}\label{sec:unc-quant}

In order build more trustworthy machine learning models, it is paramount to reliably characterize the uncertainty in their predictions \cite{smith2013uncertainty, sullivan2015introduction, soize2017uncertainty}. In the case of classification models, while the softmax scores of maximum likelihood predictions might seem to be intuitive proxies of uncertainty, it has been shown that modern neural networks do not intrinsically provide calibrated softmax scores \cite{guo2017calibration, minderer2021revisiting, bai2021don}. At best, softmax scores are heuristic notions of uncertainty. However, there exists a plethora of uncertainty quantification techniques including Bayesian methods \cite{mackay1992bayesian, neal2012bayesian, graves2011practical, blundell2015weight}, Monte Carlo dropout \cite{gal2016dropout}, and ensembles \cite{dietterich2000ensemble, lakshminarayanan2017simple, ashukha2020pitfalls}. Yet, many such methods can be limited in practice due to their computational overhead, incorrect distributional assumptions, or predisposition to specific architectures and training procedures.

\subsection{Prediction Sets and Coverage}\label{sec:sets-cov}

While many machine learning tasks are formulated as classification problems with a single ground truth label, the real world is often much more nuanced. Data points may contain elements from multiple classes, in which case multiple labels could apply. Alternatively, some data may be out-of-distribution in which case predicting any single class would misrepresent the uncertainty in the label. In such cases, forcing a predictor to only produce a single output is overly restrictive. Set-valued predictors \cite{grycko1993classification} are better equipped to deal with the ambiguities of multi-label predictions. A key desiderata of prediction sets is to construct them such that they satisfy the \emph{coverage} guarantee \cite{vovk1999machine}. Formally, we consider an input $x \in \mathcal{X} \subset \mathbb{R}^D$ with corresponding ground truth class $y \in \mathcal{Y} = \{1, \dotsc, M\}$ drawn from a joint distribution $(x, y)\sim \mathbb{P}$. Suppose we have a set-valued function $\mathcal{C}: \mathcal{X} \to 2^{[M]}$, where $[M]=\{1, \dotsc, M\}$. Prediction sets produced by $\mathcal{C}$ satisfy the coverage guarantee if\footnote{Throughout the paper, we refer to \emph{marginal} coverage since the probability is marginalized over the randomness in $(x, y)$.}
\begin{equation}\label{eq:coverage-guarantee}
     \mathbb{P}[y \in \mathcal{C}(x)] \geq 1 - \alpha,
\end{equation}
where $\alpha \in [0,1]$ is the error rate, also referred to as risk. Under the coverage guarantee, prediction sets contain the correct label with probability at least $1 - \alpha$, which is important for their reliability when presented to humans. 

{\looseness=-1}Given an arbitrary classifier $f\! : \mathcal{X} {\to}[0, 1]^M$ with softmax outputs, the simplest method of creating prediction sets is to select the \topk outputs of $f(x)$, denoted $\mathrm{top}_f(x, k)$,
\begin{equation}\label{eq:topk-set}
    \mathcal{C}_{k}(x)  = \{y \in \mathrm{top}_f(x, k) \}.
\end{equation}
While the \topk method can satisfy the coverage guarantee, it does not allow the user to specify a desired error rate \emph{a priori} \cite{chzhen2021set}. Instead, $\alpha$ must be computed as $1- \mathbb{P}[y \in \mathcal{C}_{k}(x)]$ \emph{a posteriori}. In cases where this is not directly computable, the expectation can be estimated from a calibration set $\mathcal{D}_\mathrm{cal} = \{ (x_i, y_i)\}_{i=1}^n$, with $(x_1, y_1), \dotsc, (x_n, y_n) \overset{\text{i.i.d.}}{\sim} \mathbb{P}$, obtaining the empirical risk 
\begin{equation}\label{eq:empirical-risk}
\hat \alpha = 1- \frac{1}{n}\sum_i^n \mathds{1}[y_i \in \mathcal{C}_{k}(x_i)],
\end{equation}
where $\mathds{1}$ is the indicator function. $\hat \alpha$ is a random variable over the randomness of $\mathcal{D}_\mathrm{cal}$ with expected value $\alpha$, making it an unbiased estimator of $\alpha$. 

\subsection{Conformal Prediction}\label{sec:conf}

Conformal prediction \cite{vovk2005algorithmic, shafer2008tutorial} is a prominent framework to construct prediction sets for arbitrary classification models\footnote{While conformal prediction can be applied to a broader class of discriminative models, we focus our study on classification.} such that, for any pre-specified $\alpha$, the resulting sets provably satisfy the coverage guarantee in \autoref{eq:coverage-guarantee}. In particular, it provides risk control -- the ability to choose $\alpha$ -- which is achieved by allowing the size of prediction sets to vary based on a heuristic notion of uncertainty built into $f$, such as softmax scores. First, we define a conformal score function $s\! : \mathcal{X}{\times} \mathcal{Y}{\to} \mathbb{R}$ where larger conformal scores indicate \emph{worse} agreement between an input $x \in \mathcal{X}$ and a class $y \in \mathcal{Y}$ according to the heuristic uncertainty notion. While improved alternatives exist \cite{romano2020aps, angelopoulos2021raps}, the canonical form of the conformal score for classification is $s(x,y) = 1{-}f(x)_y$, where $f(x)_y$ is the softmax output for class $y$. Second, using a calibration dataset $\mathcal{D}_\mathrm{cal}$ of size $n$ drawn from $\mathbb{P}$, we compute the conformal threshold $\hat q$ as the $\tfrac{\lceil{(n+1)(1-\alpha)}\rceil}{n}$ quantile of the conformal scores $s_i=s(x_i, y_i)$ on $\mathcal{D}_\mathrm{cal}$. Since this quantile is computed from pairs of input data $x_i$ and corresponding labels $y_i$, it calibrates the conformal score for the model.
Finally, $\hat q$ is used to generate prediction sets that contain all classes $y$ such that the conformal score is below the threshold $\hat q$,\footnote{See \autoref{fig:human_uncertainty} where $f(x)_y{>}1-\hat q$ is used in lieu of $s(x, y) {<} \hat q $.}
\begin{equation}\label{eq:conf-set}
    \mathcal{C}_{\hat q}(x)  = \{y \mid s(x, y) < \hat q \}.
\end{equation} 
Conformal prediction sets will obey the coverage guarantee in \autoref{eq:coverage-guarantee} for test data $\mathcal{D}_\mathrm{test}$ with the sole assumption being that the calibration and test data are exchangeable \cite{vovk1999machine}, a weaker assumption than being i.i.d. Hence, unlike other methods for uncertainty quantification, conformal prediction does not rely on assumptions about the underlying model $f$, its training data, nor on distributional assumptions about the data \cite{angelopoulos2022gentle}.

Importantly, conformal prediction sets are calibrated to be larger when the model has greater uncertainty in its predictions. Compared to the fixed-sized nature of \topk sets, well-designed conformal sets will have a smaller average set size for the same coverage \cite{angelopoulos2021raps}. This is a potentially useful feature for improving downstream human decision making, although one that has not been scientifically verified in prior work.

\subsection{Human-in-the-Loop Conformal Prediction}
While conformal prediction provides statistically rigorous guarantees, its applicability in real-world scenarios is limited since it requires a post-hoc mechanism to convert prediction sets into single actionable outcomes. A small number of prior works have explored using conformal prediction with human decision makers. 
\citet{straitouri2023improving} proposed restricting experts to only select an option from a conformal prediction set, and developed a search method to find an $\alpha$ that maximizes the expert's accuracy. \citet{babbar2022utility} proposed learning a policy that is attuned to a given expert, and defers automated model decisions when it expects the human is likely to be more accurate. 

These works are based on the assumption that humans can effectively make use of prediction sets compared to acting without assistance. However, such assumptions have not been supported by scientific evidence. \citet{straitouri2023improving} did not conduct any tests involving humans, instead creating algorithmic ``experts''. Follow-up work by \citet{straitouri2023designing} did involve humans, but only compared cases where humans are or are not forced to choose their answer from a conformal set. \citet{babbar2022utility} recruited humans to compare the usefulness of top-1 predictions to conformal sets and their proposed deferral scheme. However, in these tests the sets of images shown to participants were hand-selected by the experimenters to highlight cases where the top-1 prediction was incorrect, or where non-deferred sets were smaller in size. Due to such data manipulations, the test data would not be exchangeable with the calibration data, invalidating coverage guarantees. In our work we provide scientific evidence in support of the notion that humans can effectively leverage prediction sets and the uncertainty quantification that conformal prediction provides.

\section{Method}\label{sec:method}

Our aim is to determine if providing \emph{conformal} prediction sets to humans can benefit their performance on tasks via a pre-registered experiment with human subjects. We consider two aspects of performance: accuracy and speed. Prediction sets that come with coverage guarantees represent additional information, so it is natural to expect that humans could leverage that information to improve their decision accuracy. That additional information may also help them narrow their focus to a few likely candidates, or increase their cognitive load \cite{beach193broadening}, so it is plausible that predictions sets could increase \emph{or} decrease decision making speed.

Primarily, we aim to disentangle the effects of merely receiving several alternatives versus the uncertainty quantification provided by variable conformal set sizes. To this end, we ask humans to complete a challenging task, and either provide them with no assistance (control), a \topk set, or a conformal prediction set generated with the RAPS algorithm \cite{angelopoulos2021raps} (see Appendix \ref{appsub:raps}). Both types of sets are derived from the same pre-trained model and come with (empirical) coverage guarantees. To isolate the uncertainty quantification aspects of conformal prediction from the level of coverage, we ensure that both types of predictions sets achieve the same empirical coverage. First, we select a $k$ for top-$k$, and evaluate the empirical risk $\hat \alpha$ from \autoref{eq:empirical-risk} using $\mathcal{D}_\mathrm{cal}$. Then, we compute a calibrated conformal threshold $\hat q$ using the same $\mathcal{D}_\mathrm{cal}$ and with the choice of risk tolerance $\alpha$ equal to the $\hat\alpha$ achieved by top-$k$. In both cases, new data $\mathcal{D}_\mathrm{test}$ drawn i.i.d. from the same distribution as $\mathcal{D}_\mathrm{cal}$ is shown to humans along with the appropriate set and coverage guarantee.\footnote{We refrain from providing any other possible information, such as the model's ordering of likely labels, or softmax scores.}

Our experiments are forced-choice tasks, where participants are shown a stimulus $x$ and must select one label $y$ out of all possible classes. Each participant is randomly assigned to one task and one treatment (control, top-$k$, or conformal), then is trained on their task, and finally completes $m$ trials. We collect data on their answers and response times.

The tasks we employ have one-hot labels, so that correctness is binary. Accuracy is computed as the fraction of the $m$ trials that were answered correctly. For each of $N$ participants in a treatment, we consider their accuracy over $m$ trials as one observation; for large $N$ we expect the mean accuracy follows a Gaussian distribution by the central limit theorem. Also, since each participant takes only one test, observations are independent and unpaired. Hence, we can perform Welch's $t$-test \cite{welch1947generalization} to test the null hypothesis that two treatments have the same mean accuracy.\footnote{For accuracy, we perform one-sided $t$-tests with the prior expectation that top-$k$ sets are more helpful than the control, and that conformal sets are more helpful than \topk sets due to their smaller average size.}

For measurement of decision making speed, we capture the time between presentation of a stimulus and when the participant enters their response. For every response, participants are unconstrained in the amount of time they spend deciding on their answer, but have a monetary incentive to complete the test expediently since compensation is primarily a fixed amount as enforced by the recruitment platform \cite{prolific}. We treat each participant's total response time over $m$ trials as a single observation, and again apply Welch's $t$-test for the null hypothesis that two treatments have the same mean response time.

Finally, we provide effect sizes computed with Cohen's $d$ for equal sample sizes: $d = (\overline{X}_1 - \overline{ X}_2)/{\sqrt{(s_1^2 + s_2^2) / 2}}$. Here, $\overline{X}_i$ is the sample mean over observations for treatment $i$, and the sample variance is $s_i^2=\frac{1}{N-1}\sum_j^{N}(X^j_i - \overline{X}_i)^2$, using observations $X^j_i$ from treatment $i$ \cite{cohen2013statistical}.

\section{Experiments and Evaluation}\label{sec:experiments}

\subsection{Tasks, Datasets, and Models}

We designed three independent tasks to represent a variety of real-world settings where human decision makers might benefit from model assistance. Each task is based on a dataset from the machine learning literature for which pre-trained models are available. As discussed in Section \ref{sec:conf}, it is not required that the models were trained on the datasets we employ. The three tasks do not rely on specialized knowledge, only fluency in English, so that they can be completed by the general population. 
Full details on dataset construction are given in \autoref{app:implementation}, and our code is available at \href{https://github.com/layer6ai-labs/hitl-conformal-prediction}{this GitHub repository} for reproducibility.

\vspace{-10pt}
\paragraph{Image Classification}
Image classification, a mainstay application of computer vision, is widely performed by humans in critical settings every day. For instance, radiologists diagnose diseases by classifying X-ray images, and could potentially improve the accuracy and speed of diagnosis by leveraging machine learning \cite{choy2018radiology, agarwal2023combining}. As a representative image classification task we used ObjectNet \cite{barbu2019objectnet}, a dataset of common objects photographed from intentionally difficult viewpoints. We selected the 20 most common classes to reduce the number to a manageable level for humans, balanced the selected classes using stratified sampling, then split the dataset into $\mathcal{D}_{\mathrm{cal}}$ and $\mathcal{D}_{\mathrm{test}}$ maintaining class balance. We used CLIP ViT-L/14 \cite{radford2021clip} as a zero-shot classifier. 

\vspace{-10pt}
\paragraph{Sentiment Analysis}

Sentiment analysis is crucial in human interactions. For instance, social media users routinely classify sentiments in posts, comments, and reviews to understand context and communicate effectively. As a prototypical task, we used GoEmotions \cite{demszky2020goemotions}, a dataset of Reddit comments in English with annotations of sentiment categories. 
We selected the 10 most common classes based on the validation dataset, and balanced the classes using stratified sampling for the validation and test set separately to form $\mathcal{D}_{\mathrm{cal}}$ and $\mathcal{D}_{\mathrm{test}}$. For the model we selected a RoBERTa-Base \cite{liu2019roberta} fine-tuned on the GoEmotions training set \cite{lowe2023goemotionsmodel}.

\vspace{-10pt}
\paragraph{Named Entity Recognition (NER)}
Humans regularly perform NER in daily life as we encounter unfamiliar proper names and use context to infer what type of entity that name represents. In the financial industry, human annotators are often employed to locate and extract entities from legal documents, but may rely on model recommendations to expedite extraction. We used the Few-NERD dataset of sentences from Wikipedia in English \cite{ding2021fewnerd}, with each word annotated as a named entity class. For our task, a single entity was selected from every sentence as the classification target. 
We selected the 10 most common classes based on the validation dataset, and balanced the classes using stratified sampling for the validation and test set separately to form $\mathcal{D}_{\mathrm{cal}}$ and $\mathcal{D}_{\mathrm{test}}$. 
For the model we used a SpanMarker RoBERTa-Large fine-tuned on the Few-NERD training set \cite{aarsen2023spanmarker, aarsen2023spanmarkermodel}.

For all three tasks we used $k=3$ to compare the usefulness of top-$k$ sets and conformal sets.
A summary of model performance as percentages is given in \autoref{tab:model-performance}. We show the empirical risk $\hat \alpha$ for top-$k$ sets computed on $\mathcal{D}_{\mathrm{cal}}$, which is then used as $\alpha$ for conformal calibration. We also show for $\mathcal{D}_{\mathrm{test}}$ the top-1 and top-3 accuracy of the models, empirical conformal set coverage, and average conformal set size.

\begin{table}[t]
    \centering
    \setlength{\tabcolsep}{3pt}
    \caption{Model Performance}
    \label{tab:model-performance}
    \vspace{-8pt}
    \begin{tabular}{lccccc} \toprule
        Task & $\hat \alpha$ & Top-1 &  Top-3 & Coverage & Avg. Size \\ \hline
        ObjectNet & 0.065 & 83.3 & 95.0 & 94.1 & 1.68\\ \hline
        GoEmotions & 0.085 & 67.2 & 94.4 & 91.8 & 2.49\\ \hline
        Few-NERD & 0.021 & 91.1 & 98.3 & 98.2 & 1.82 \\ \bottomrule
    \end{tabular}
    \vspace{-13pt}
\end{table}

\vspace{-2pt}
\subsection{Experiment Design}
\vspace{-2pt}
We created our human subject experiments using PsychoPy \cite{peirce2019psychopy} and hosted them on Pavlovia \cite{pavlovia}. Participants were recruited on Prolific \cite{prolific}, a platform that provides the highest quality data according to scientific comparisons of behavioural data collection platforms \cite{eyal2021dataquality, douglas2023dataquality}.

Our experimental design was informed by research on the collection of high quality data from crowd sourcing platforms. \citet{mitra2015comparing} show that the most reliable way to ensure high quality data is by training participants on the task at hand, while providing financial incentives can also be beneficial. Hence, in each experiment participants were asked for consent, given instructions on the task, and trained with 20 examples after which the testing phase began. Participants were paid a fixed amount with a financial incentive proportional to their accuracy. On average, participants were paid 7.80 GBP/hr, and in total we spent 1500 GBP on participant compensation, excluding fees.

\begin{figure}[t]
\centering
\includegraphics[width=0.9\columnwidth, trim={50, 0, 100, 0}, clip]{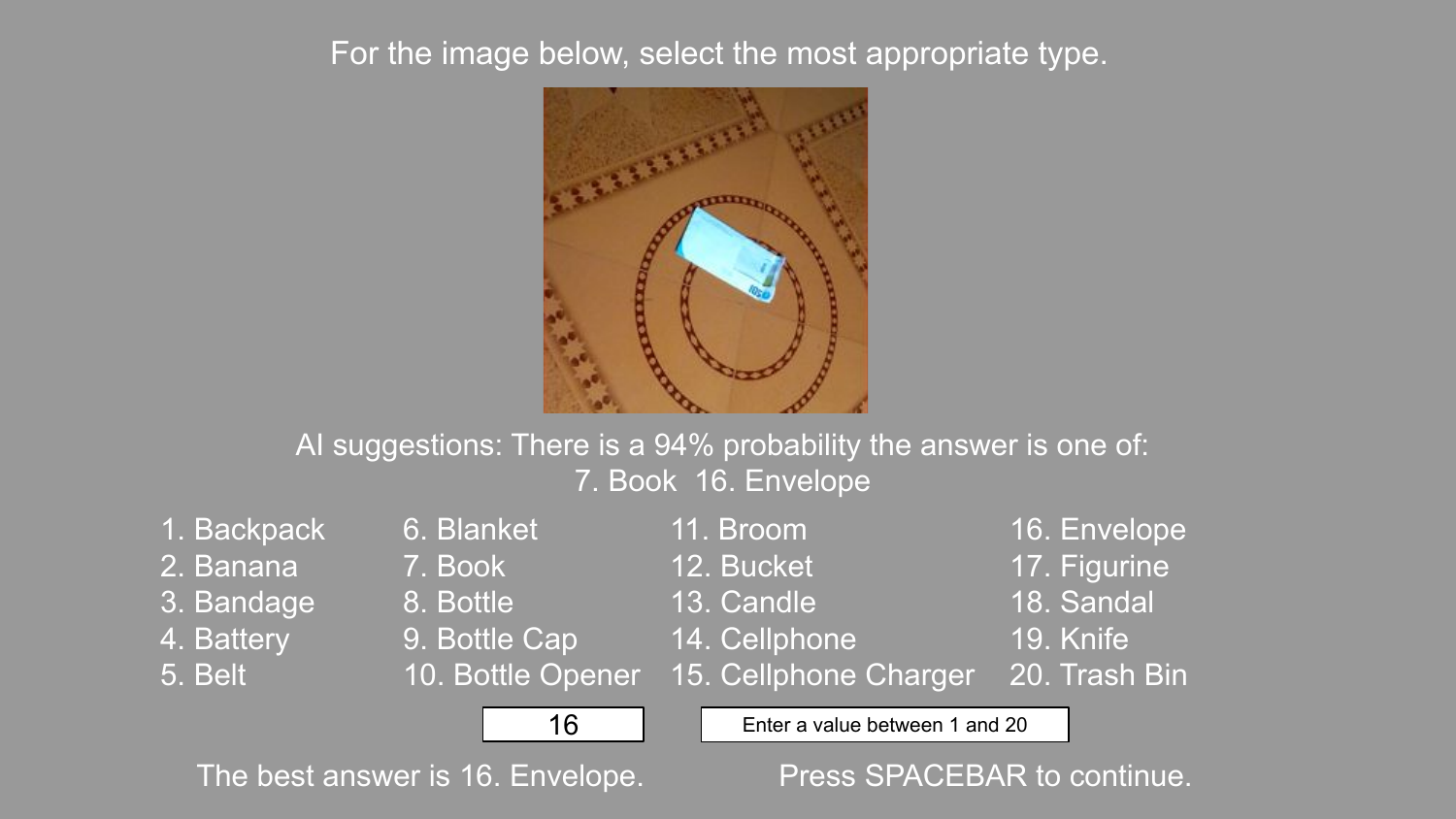}
\vspace{-3pt}
\caption{Main trial screen shown to participants for ObjectNet with conformal set treatment. The correct answer is given only after the participant responds.}
\label{fig:objectnet-trial}
\vspace{-5pt}
\end{figure}

An example trial screen is shown in \autoref{fig:objectnet-trial}, while complete descriptions of the tasks are given in \autoref{app:experiment}. On each trial the participant was shown a stimulus $x$ and all $M$ class labels, and was forced to classify $x$ as one of the labels. For the \topk and conformal treatments, they were also shown a prediction set, along with the (empirical) coverage guarantee. The correct answer was displayed once the participant confirmed their decision. There was no time limit to enter responses, although for ObjectNet the stimulus was only shown for 0.22s, within the limits of human perception \cite{fraisse1984perception}, to increase the difficulty of the task. The set of $m$ stimuli shown to each participant was drawn from $\mathcal{D}_{\mathrm{test}}$, making each test randomized.

\begin{figure}[t]
\centering
\includegraphics[width=1\columnwidth, trim={0, 20, 0, 30}, clip]{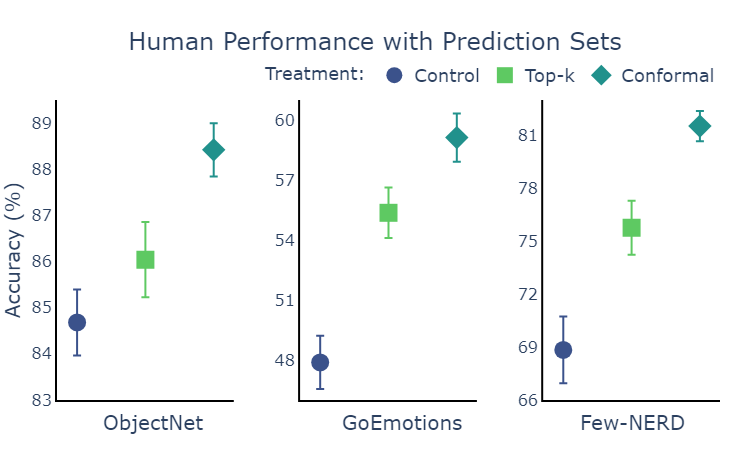}
\caption{Human performance (accuracy) across three tasks and three treatments. Data is shown as mean accuracy, while error bars show unbiased standard errors ($N=50$).}
\label{fig:main-acc}
\vspace{-7pt}
\end{figure}

\begin{figure}[h!]
\centering
\includegraphics[width=1\columnwidth, trim={0, 20, 0, 30}, clip]{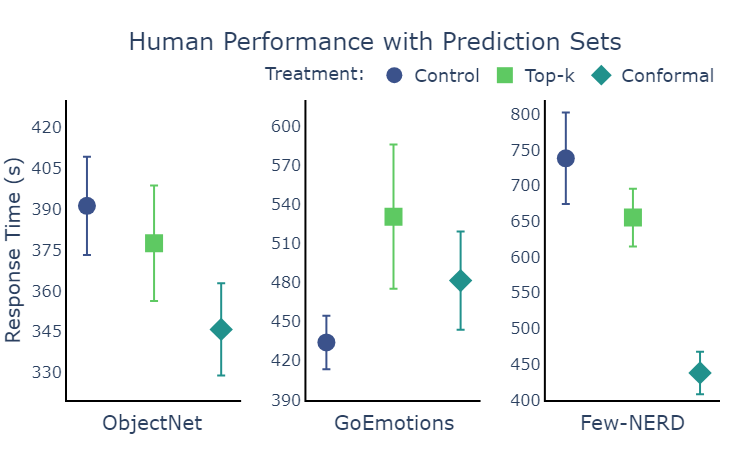}
\caption{Human performance (speed) across three tasks and three treatments. Data is shown as mean response time (s), while error bars show unbiased standard errors ($N=50$).}
\label{fig:main-time}
\vspace{-10pt}
\end{figure}

\section{Results}\label{sec:results}

\subsection{Human Performance Measurement}\label{subsec:main-results}

Our main experiment covered three tasks and three treatments, for which we recruited $N=50$ unique people each, totalling 450 paid participants. The results for mean accuracy across observations are shown in \autoref{fig:main-acc}, while \autoref{fig:main-time} shows the average time taken for participants to complete all test trials.

\begin{table}[t]
    \centering
    \setlength{\tabcolsep}{5pt}
    \caption{Accuracy -- $p$-values and Effect Sizes}
    \label{tab:p-values-acc}
    \small
    \begin{tabular}{lcccccc} \toprule
        & \multicolumn{2}{c}{ObjectNet} & \multicolumn{2}{c}{GoEmotions} &  \multicolumn{2}{c}{Few-NERD} \\
        \cmidrule(lr){2-3} \cmidrule(lr){4-5} \cmidrule(lr){6-7}
        Comparison & $p$ & $d$ & $p$ & $d$ & $p$ & $d$ \\ \hline
        Top-$k$ $>$ Control & \textcolor{red}{$0.1$} & 0.3 & $5\mathrm{e-}{5}$ & 0.8 & $0.003$ & 0.6 \\ \hline
        Conf. $>$ Control & $5\mathrm{e-}{5}$ & 0.8 & $5\mathrm{e-}{9}$ & 1.0 & $3\mathrm{e-}{8}$ & 1.0 \\ \hline
        Conf. $>$ Top-$k$ & $0.01$ & 0.5 & $0.02$ & 0.4 & $8\mathrm{e-}{4}$ & 0.7\\ \bottomrule
    \end{tabular}
\end{table}

\begin{table}[h!]
    \centering
    \setlength{\tabcolsep}{5pt}
    \caption{Response Time -- $p$-values and Effect Sizes}
    \label{tab:p-values-time}
    \small
    \begin{tabular}{lcccccc} \toprule
        & \multicolumn{2}{c}{ObjectNet} & \multicolumn{2}{c}{GoEmotions} &  \multicolumn{2}{c}{Few-NERD} \\
        \cmidrule(lr){2-3} \cmidrule(lr){4-5} \cmidrule(lr){6-7}
        Comparison & $p$ & $d$ & $p$ & $d$ & $p$ & $d$ \\ \hline
        Top-$k$ vs. Control & \textcolor{red}{$0.6$} & 0.1 & \textcolor{red}{$0.1$} & 0.3 & \textcolor{red}{$0.3$} & 0.2 \\ \hline
        Conf. vs. Control & \textcolor{red}{$0.07$} & 0.4 &\textcolor{red}{$0.3$} & 0.2 & $6\mathrm{e-}{5}$ & 0.9 \\ \hline
        Conf. vs. Top-$k$ & \textcolor{red}{$0.2$} & 0.2 & \textcolor{red}{$0.5$} & 0.1 & $4\mathrm{e-}{5}$ & 0.9 \\ \bottomrule
    \end{tabular}
\vspace{-10pt}
\end{table}

For accuracy, we see that \topk sets mostly do improve human performance compared to the control, while conformal sets lead to better performance than \topk sets. To confirm these visual trends, we conducted pre-planned comparisons between pairs of treatments using $t$-tests as described in Section \ref{sec:method}. In all cases except one we reject the null hypothesis that the mean accuracy is the same between treatments, using the significance threshold $p<0.05$ (\autoref{tab:p-values-acc}, non-significant results in red). Notably, for all three independent tasks we find statistically significant evidence that conformal prediction sets are more useful to humans than \topk sets, with medium effect sizes ($0.4\leq d\leq 0.7$)  \cite{cohen2013statistical}. Because both types of prediction sets have the same (empirical) coverage, there are only two differences between the methods to which we can ascribe the improvement: conformal sets are smaller on average (\autoref{tab:model-performance}), and conformal sets quantify uncertainty.

We also pre-planned comparisons between the treatments on the average response time. However, in viewing \autoref{fig:main-time} we do not see a consistent trend between all tasks. Conformal sets have the lowest average response time on two tasks, while the control treatment led to the fastest completions on GoEmotions. In most cases we do not reject the null hypothesis that the treatment has no effect on the mean response time (\autoref{tab:p-values-time}). The additional information provided as a prediction set must be processed by the user and incorporated into decision making, which can outweigh the speed advantages of receiving a curated shortlist.

The remainder of our analysis was not explicitly planned as part of our pre-registration, so is presented for insight without statistical analysis or claims of significance. Additional analysis of our data is given in Appendix \ref{app:analysis}.

\vspace{-2pt}
\subsection{Ablations}\label{subsec:ablations}
\vspace{-2pt}
Based on the results of our pre-registered experiment in Section \ref{subsec:main-results}, we conducted targeted ablations that independently varied two aspects of the conformal prediction framework, namely conformal set sizes and model performance. For GoEmotions we fixed the model, and performed conformal prediction with a less optimized RAPS procedure that produced larger average set sizes, while for ObjectNet we swapped out the CLIP ViT-L/14 for a weaker CLIP ViT-B/32. We re-ran three experiments (GoEmotions conformal, and ObjectNet top-$k$ and conformal) for which we recruited an additional $N=50$ unique people each, or 150 in total.

\begin{figure}[t]
    \centering
    \includegraphics[width=\columnwidth, trim={0, 0, 50, 0}, clip]{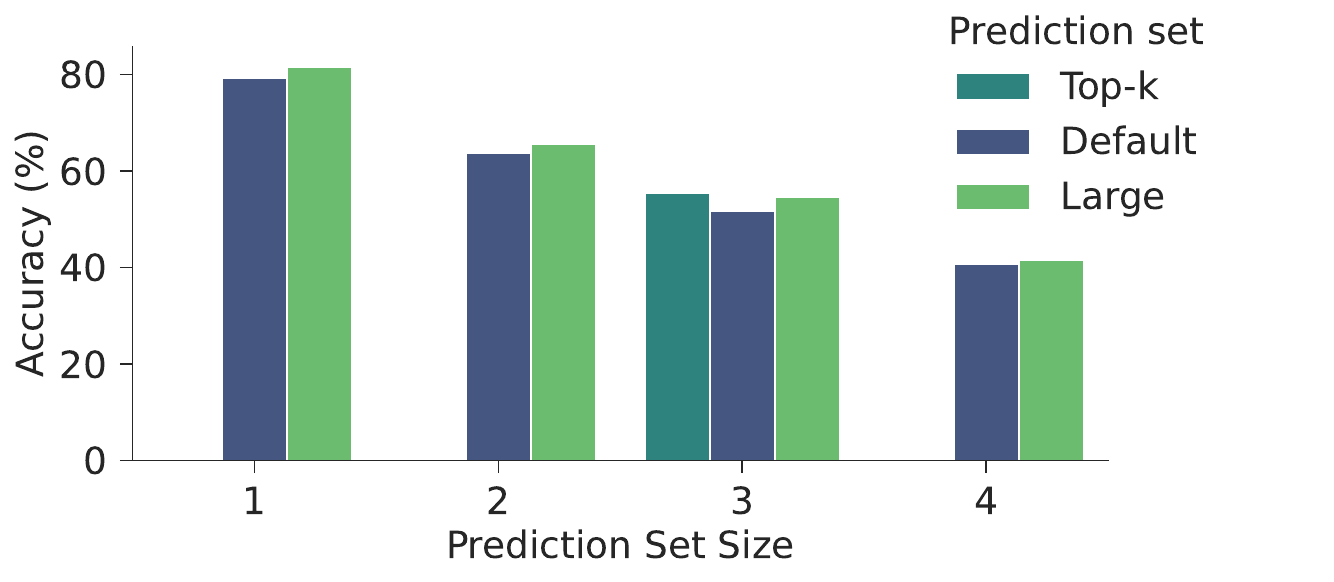}
    \caption{Accuracy by prediction set size on GoEmotions.}
    \label{fig:go-ablation-acc}
\end{figure}
\begin{figure}[t]
    \centering
    \includegraphics[width=\columnwidth, trim={0, 0, 10, 0}, clip]{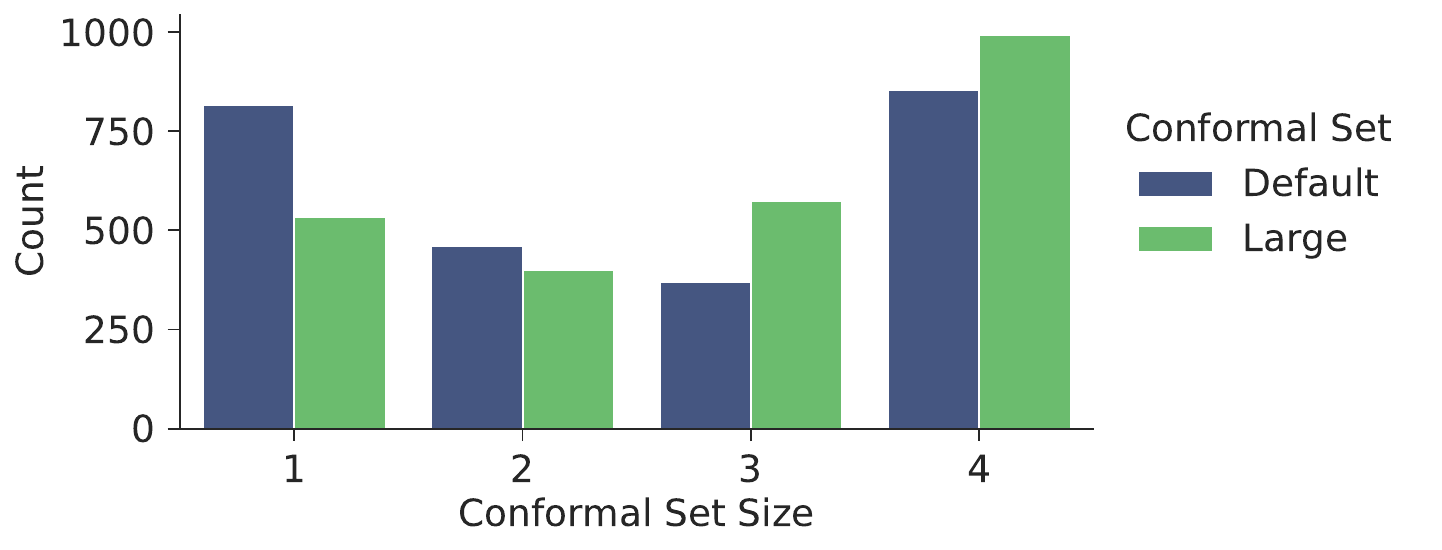}
    \caption{Histograms of conformal set size on GoEmotions.}
    \label{fig:go-ablation-count}
\vspace{-10pt}
\end{figure}

\begin{figure*}[t]
    \centering
    \subcaptionbox{Human performance\label{fig:obj-ablation-acc}}{\includegraphics[width=0.24\linewidth]{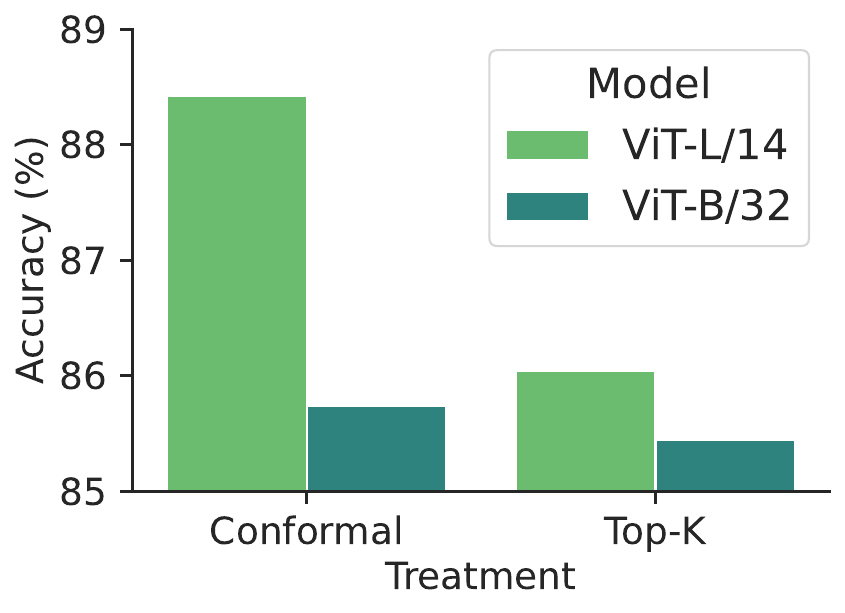}}
    \subcaptionbox{Empirical coverage\label{fig:obj-ablation-cvg}}{\includegraphics[width=0.24\linewidth]{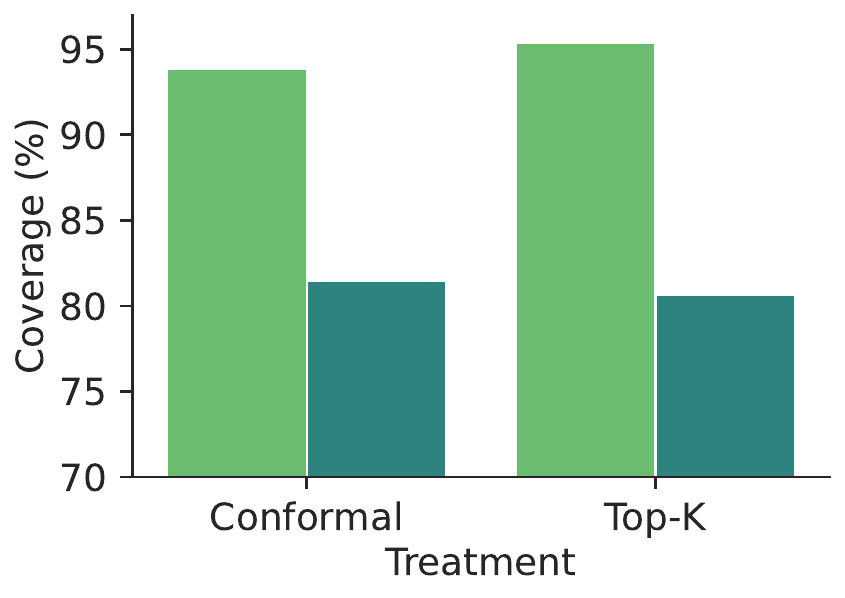}}
    \subcaptionbox{Adoption rate\label{fig:obj-ablation-adoption}}{\includegraphics[width=0.24\linewidth]{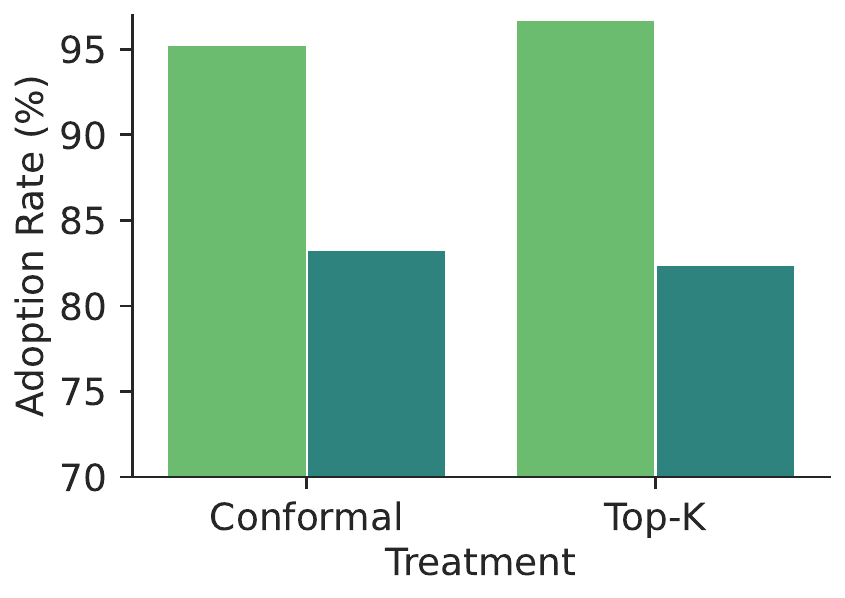}}
    \subcaptionbox{Conformal set size\label{fig:obj-ablation-conf-size}}{\includegraphics[width=0.24\linewidth]{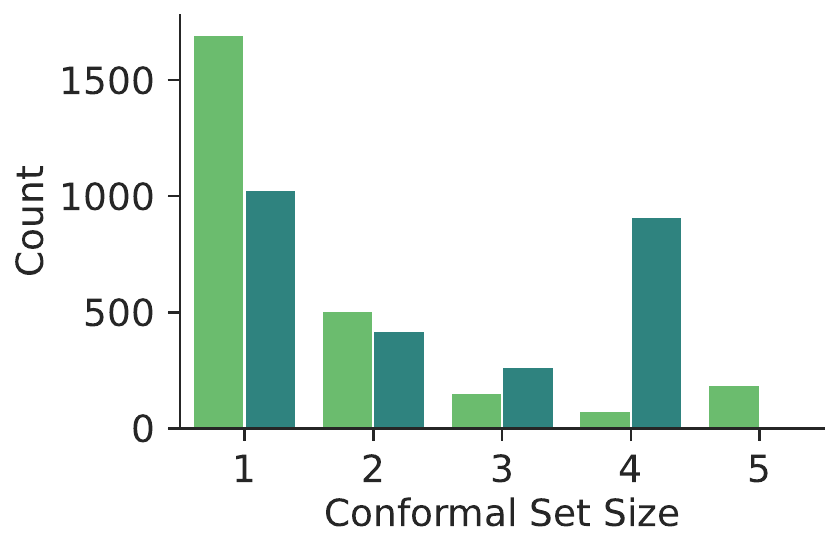}}
    \caption{
    Comparison of ObjectNet results between two models. The ViT-L/14 model achieved 83.3\% top-1 accuracy, which was significantly better than ViT-B/32 at 61.1\% on ObjectNet test data $\mathcal{D}_{\mathrm{test}}$.
    }
    \label{fig:obj-ablation}
    \vspace{-10pt}
\end{figure*}

\vspace{-10pt}
\paragraph{Conformal Set Size}
We explored the effect of conformal set size on human accuracy with GoEmotions by comparing two settings: the ``default'' setting used in our main experiment with an average conformal set size of 2.49 class labels, and the ``large'' setting with an average size of 2.77. With larger conformal sets, humans achieved lower accuracy, $56.9\pm1.3$\% compared to $59.1\pm1.2$\% with the default setting (cf. $55.4\pm1.3$\% for top-3 sets). This is consistent with the intuitive notion that smaller average set sizes should be more helpful for the same coverage guarantee.

Examining the set size distribution between the two settings helps reveal the source of performance differences. As shown in \autoref{fig:go-ablation-acc}, human performance decreases on examples where prediction sets were larger, but importantly human performance conditioned on set size was comparable between the settings. By comparison, \autoref{fig:go-ablation-count} shows that the large setting generated fewer singleton sets where human accuracy is highest, resulting in the observed overall decrease in human accuracy. Optimizing for set sizes where human performance is highest is one way to improve the usefulness of conformal sets for humans.

\vspace{-10pt}
\paragraph{Model Accuracy}
By swapping the ViT-L/14 model used in Section \ref{subsec:main-results} for a weaker ViT-B/32, we can study the effect of model performance on the usefulness of both \topk and conformal prediction sets. \autoref{fig:obj-ablation-acc} shows conformal sets greatly enhanced human performance when using a superior underlying model, whereas for \topk the improved model did not translate to as much human improvement. This is despite the empirical coverage (\autoref{fig:obj-ablation-cvg}) increasing by a similar amount between the treatments (by design of our calibration). Furthermore, the rate at which humans chose an answer from the prediction set, which we call the adoption rate (\autoref{fig:obj-ablation-adoption}), was consistently close to the empirical coverage (coverage was communicated to participants during the test). The outsized increase in accuracy for conformal when the better model was used is a function of the higher quality of sets that it produces; although coverage and adoption rates are the same, humans must still choose the correct answer out of the set when they adopt from it, which is much easier to do for singleton sets. The stronger model produces a distribution of conformal set sizes more skewed towards singleton sets (\autoref{fig:obj-ablation-conf-size}), whereas \topk sets always use $k=3$ even with a more confident and correct model.

\subsection{Insights}
\begin{figure}[t]
    \centering
    \includegraphics[width=1\columnwidth, trim={0, 0, 0, 0}, clip]{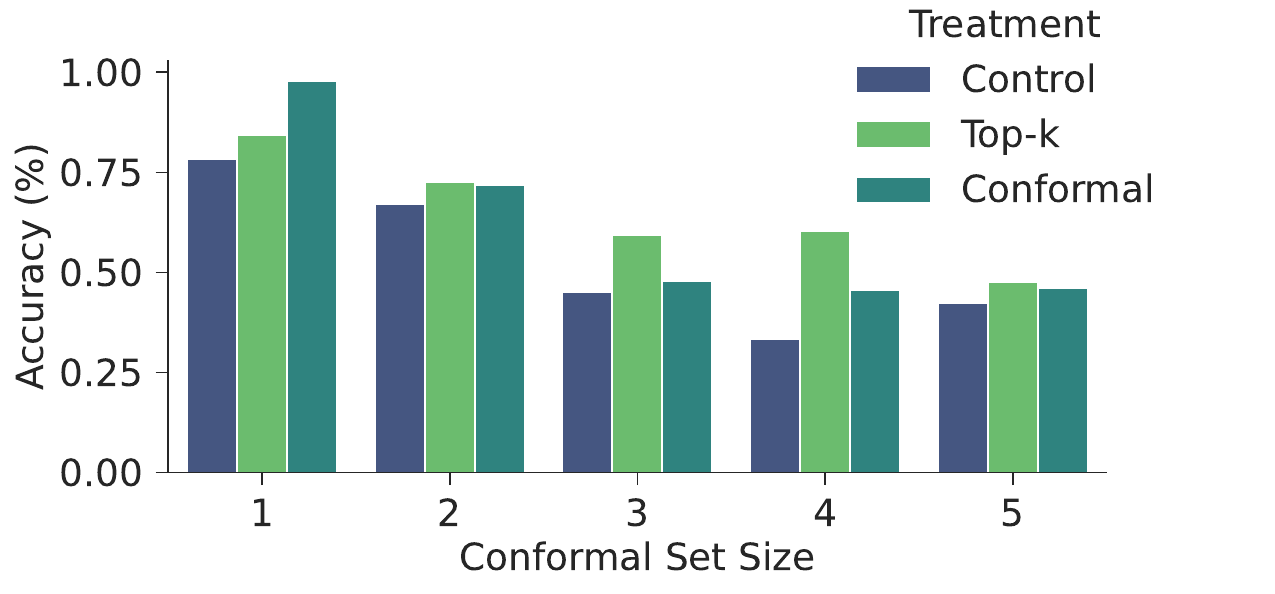}
    \caption{Human accuracy by difficulty of examples (conformal set size) on Few-NERD.}
    \label{fig:acc-set-size}
        \vspace{-10pt}
\end{figure}

\vspace{-2pt}
\paragraph{Role of Uncertainty} Within each task, some stimuli are more challenging than others which necessitates quantifying model uncertainty, in our case with the size of the conformal set. Since only the conformal treatment distinguishes examples by their difficulty, we can investigate the role of uncertainty quantification by comparing human performance across treatments conditional on conformal set size. In \autoref{fig:acc-set-size} we show this for accuracy using Few-NERD for illustration, although the trends are consistent with the other two tasks (see Appendix \ref{app:analysis}), and in \autoref{fig:time-set-size} we show the same for response times. Overall, we observe that conformal sets improve human decision making most compared to other treatments when the model expresses certainty through singleton conformal sets.

\begin{figure}[ht]
    \centering
    \includegraphics[width=1\columnwidth, trim={0, 0, 0, 0}, clip]{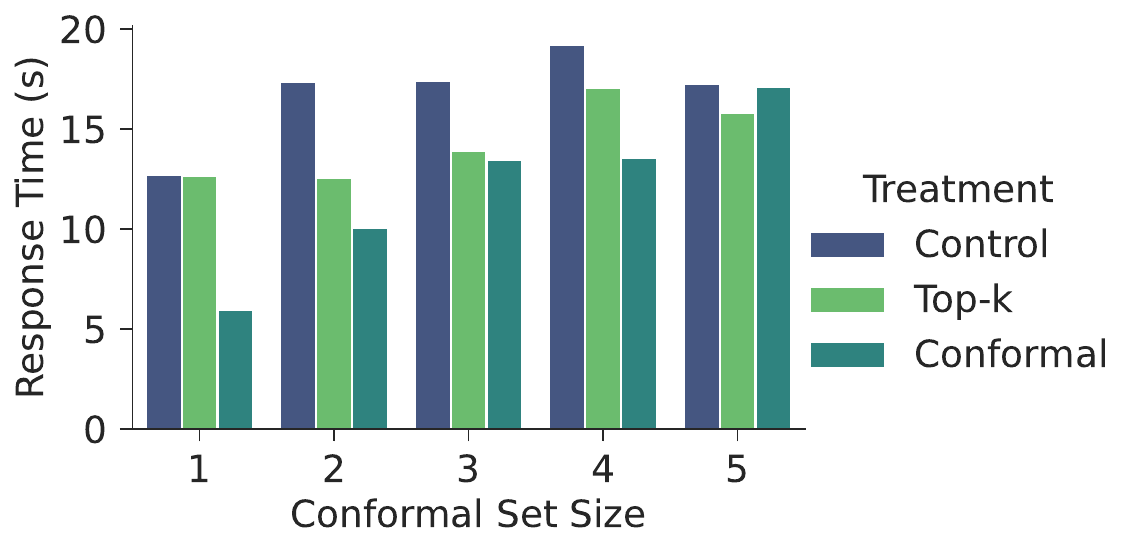}
    \caption{Human response time by difficulty of examples (conformal set size) on Few-NERD.}
    \label{fig:time-set-size}
        \vspace{-10pt}
\end{figure}

\begin{figure*}[t]
    \centering
    \includegraphics[width=\textwidth, trim={7, 10, 28, 10}, clip]{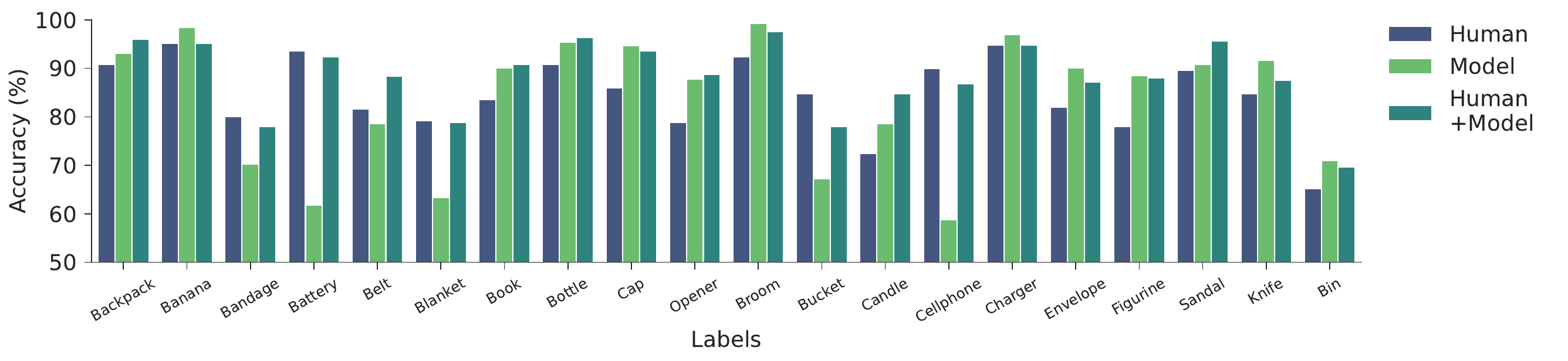}
    \caption{Per-class accuracy on ObjectNet of Human-only (control), Model-only (top-1 accuracy), and Human-Model teams (conformal).}
    \label{fig:obj-acc-by-class}
    \vspace{-14pt}
\end{figure*}

\vspace{-2pt}
Based on the trends for the control treatment we see that humans and the model are aligned in which examples they find difficult, with larger set sizes correlating with worse human performance. This suggests that conformal prediction could be leveraged to identify samples that would be challenging for humans, allowing them to optimize their efforts by allocating more attention to these examples. In doing so, practitioners should be aware that the marginal coverage guarantee may not extend to sub-populations, such as those examples assigned large prediction sets.

\paragraph{Ensembling Effects} In \autoref{fig:main-acc} we concluded that the combination of human and model (conformal treatment) outperforms humans alone (control) in terms of accuracy. Another relevant question is whether human-model teams outperform the model alone. From \autoref{tab:model-performance} we see that all treatments outperformed the top-1 accuracy of the model on ObjectNet, but the reverse is true on GoEmotions and Few-NERD. In \autoref{fig:obj-acc-by-class} we examine this comparison in more detail using per-class accuracies for ObjectNet. We see high variance in the model's per-class accuracy, possibly due to the zero-shot nature of predictions we implemented, whereas humans were more consistent (standard deviation $13.4$ vs. $7.8$). Notably, low model accuracy (Bandage, Battery, Blanket, Bucket, Cellphone) tended to drag down the human-model team compared to humans alone. This is a point of caution for implementing human-in-the-loop systems, as poor model performance or biases against certain groups may not be completely overcome by humans. However, when the model outperformed the humans, the human-model team also tended to benefit, sometimes surpassing both individual partners (Backpack, Book, Bottle, Opener, Candle, Sandal). Hence there is a type of ensembling effect \cite{dietterich2000ensemble} where one partner's weaknesses can be corrected by the other.



\vspace{-5pt}
\paragraph{Adoption Rate} Participants in the \topk and conformal treatment groups were informed that the prediction sets they were shown contained the true answer with probability $1-\hat\alpha$, but were not constrained in how they chose their answers. As shown in \autoref{tab:select-answer-from-set}, participants given conformal sets chose their answer from the set at an adoption rate very similar to the set's stated coverage guarantee, whereas reliance on \topk sets was higher than expected. This may suggest that the distributional information of variable-sized conformal sets better communicates when sets can be trusted.

\vspace{-5pt}
\begin{table}[ht]
    \centering
    \setlength{\tabcolsep}{3pt}
    \caption{Adoption Rate (\%) by Treatment Group}
    \vspace{-2pt}
    \label{tab:select-answer-from-set}
    \begin{tabular}{lr|rr} \toprule
        Task & $1-\hat \alpha$ & Top-$k$ & Conformal  \\ \hline
        ObjectNet & 94 & 97 & 95 \\ \hline
        GoEmotions & 92 & 94 & 92 \\ \hline
         Few-NERD & 98 & 99 & 99 \\ \bottomrule
    \end{tabular}
\vspace{-15pt}
\end{table}


\section{Conclusion and Outlook}

It is easy to assume without further reflection that uncertainty quantification through conformal prediction will make model predictions more useful to humans. However, in science we must not blindly follow such assumptions without evidence to support them. In this work, we conducted a scientific study to verify the intuitive notion that providing prediction sets from models can improve human decision making. By recruiting 600 paid participants across all tests, and collecting a total of 42,500 individual responses, we found statistically significant evidence that conformal prediction sets can improve human accuracy on classification tasks, both compared to no assistance, and to \topk sets. Since we controlled for the level of empirical coverage provided by conformal and \topk sets, we ascribe the improvements in accuracy to two factors: conformal sets have smaller average set sizes, and they quantify model uncertainty via their distribution of set sizes. In contrast to accuracy, we did not find consistent evidence that conformal sets increase decision making speed. A possible explanation is that parsing prediction sets can increase the amount of cognitive processing required to make an informed decision.


Our work informs the design of human-in-the-loop decision pipelines \cite{wu2022survey}, which we point out are a natural solution to the problem that set-valued predictions may not be actionable. Including a human into decision pipelines can mitigate some of the concerns around the trustworthiness of machine learning models. For example, more advanced machine learning methods are often less explainable \cite{linardatos2021explainability}, whereas humans could articulate their thought process even when aided by prediction sets. 
Neural networks are infamously susceptible to adversarial examples \citep{biggio2013evasion, szegedy2014intriguing, goodfellow2015explaining}, which, by definition, do not fool humans, making a human-in-the-loop perhaps the most robust defense to adversarial attacks. Still, there are limitations to human-in-the-loop systems. On two out of three tasks, top-1 model accuracy was higher than what humans achieved (even with model assistance). We also found that when a model performs particularly poorly on groups within the data, prediction sets can drag down human performance on those groups. This could manifest as a transfer of biases from models to humans, and reinforces that the fairness of models still needs to be considered even with a human-in-the-loop \cite{mehrabi2021fairness}.

Finally, while we only covered classification tasks in this study, it would be interesting to examine the compatibility of humans with conformal prediction for regression and time series problems. We also only considered tasks that could be completed by the general population. Our society still relies on experts, such as medical doctors, for many critical decisions and is not prepared to accept the risks of automating them. However, there are potentially great benefits to augmenting the skills of experts by providing information from a model in the form of conformal prediction sets, leaving ultimate control in the hands of the experts \cite{tizhoosh2018medical, kompa2021second}.
\clearpage
\section*{Impact Statement}
Our work promotes the integration of humans into decision making pipelines as a means to mitigate negative impacts of untrustworthy machine learning. As such, we do not foresee negative societal impacts of our study. We provide a complete description of research ethics for our human subject experiments in Appendix \ref{app:experiment}.

\section*{Acknowledgements}
We would like to thank Mouloud Belbahri and Brendan Ross for comments on a draft of this manuscript.

\bibliography{references}
\bibliographystyle{icml2024}

\newpage
\appendix
\onecolumn
\section{Implementation Details}\label{app:implementation}

\subsection{Dataset Pre-processing}\label{appsub:datasets}
High level statistics about the calibration and test datasets are given in \autoref{tab:dataset-description}. Below, we give complete details on the pre-processing applied to each dataset. All pre-processing including calibration and generation of conformal sets was carried out with an Intel Xeon Silver 4114 CPU and TITAN V GPU and takes under 3 hours. We provide the code to generate all datasets at \href{https://github.com/layer6ai-labs/hitl-conformal-prediction}{github.com/layer6ai-labs/hitl-conformal-prediction}.

\begin{table}[ht]
    \centering
    \caption{Dataset Statistics}
    \label{tab:dataset-description}
    \begin{tabular}{ccccc} \toprule
        Dataset & $| \mathcal{D}_{\mathrm{cal}} |$  & $| \mathcal{D}_{\mathrm{test}} |$ &  Total Classes & Used Classes ($M$)\\ \hline
        ObjectNet & 2000 & 2620 & 313 & 20 \\\hline
        GoEmotions & 1180 & 1030 & 28 & 10 \\ \hline
        Few-NERD & 5000 & 2000 & 66 & 20 \\ \bottomrule
    \end{tabular}
    \vspace{-10pt}
\end{table}

\paragraph{ObjectNet} Since ObjectNet is comprised of only a test set (50,000 images), we split it to obtain images for calibration. First, we selected the $M=20$ most common classes from all 313 to ensure humans could easily parse all the listed classes. The resulting classes were: [\textit{`Backpack', `Banana', `Bandage', `Battery', `Belt', `Blanket', `Book', `Bottle', `Bottle Cap', `Bottle Opener', `Broom', `Bucket', `Candle', `Cellphone', `Cellphone Charger', `Envelope', `Figurine', `Sandal', `Knife', `Trash Bin'}]. We then separated the selected samples into calibration and test datasets, reserving 2000 samples for calibration with the remaining samples used for testing. During splitting, we additionally performed stratified sampling on both the calibration and test sets separately to ensure that classes were balanced. As both calibration and test sets were treated the same way, they can be considered i.i.d. for conformal prediction. Finally, each image was resized via bicubic and center-cropping transformations to $224 \times 224$ pixels following the image processing pipeline of CLIP \cite{radford2021clip} so that both the model and humans were presented with consistent images. ObjectNet is released under a license that permits free use for research purposes.

\paragraph{GoEmotions} We used the original validation dataset for conformal calibration and the original test set as our test set. For pre-processing, we kept only the sentences with a single label and removed any sentences containing emojis, as emojis were incompatible with PsychoPy. Then, to make the task more manageable for human evaluators, we picked the $M=10$ most common sentiment classes based on the calibration set [\textit{`Admiration', `Gratitude', `Approval', `Disapproval', `Amusement', `Annoyance', `Curiosity', `Love', `Optimism', `Neutral'}] out of the 28 overall classes. Finally we performed stratified sampling on both calibration and test sets so that the selected classes were balanced. As both calibration and test sets were treated the same way, they can be considered i.i.d. for conformal prediction. The resulting datasets contained 1180 calibration samples and 1030 test samples. GoEmotions is released under a license that permits free use for research purposes.

\paragraph{Few-NERD} We used the validation and test sets that are defined in the supervised variant of Few-NERD as our calibration and test sets, respectively. We first filtered out text examples that contained non-ASCII characters because of limitations when displaying through PsychoPy. We also filtered out examples where the longest named entity spans more than 8 words, where the entire example is longer than 60 words, and where the named entity is presented without any context. Then we selected the $M=20$ most common classes which were [\textit{`person-actor', `person-artist/author', `person-athlete', `other-award', `location-bodiesofwater', `other-biologything', `organization-company', `event-attack/battle/war/militaryconflict', `organization-education', `location-GPE', `organization-government/governmentagency', `organization-media/newspaper', `art-music', `organization-politicalparty', `person-politician', `event-sportsevent', `organization-sportsleague', `organization-sportsteam', `location-road/railway/highway/transit', `art-writtenart'}]. For presentation to humans we relabeled these classes as [\textit{`Actor/Actress', `Artist/Author', `Athlete', `Award', `Body of Water', `Biology', `Company', `Military Conflict', `Education', `Geopolitics', `Government', `Media Organization', `Music', `Political Party', `Politician/Leader', `Sports Event', `Sports League', `Sports Team', `Transportation Route', `Written Art'}]. We performed stratified sampling separately on the validation and test datasets to balance their classes such that a single named entity was selected and highlighted for examples that contained multiple. Hence this became a balanced multi-class classification task. Finally, we reduced the size of the test set because only 2,500 individual trials were planned for each treatment ($N=50$ participants, $m=50$ trials each). Throughout the process the calibration and test sets were treated the same way, so we can consider them i.i.d. for conformal prediction. The final datasets comprised 5000 calibration samples and 2000 test samples. Few-NERD is released under a license that permits free use for research purposes.

In our pre-registration, we stated that one of the three datasets used would be Broad Twitter Corpus \cite{derczynski2016broad}, an NER task gathered from Twitter data. However, upon preparing this dataset for experiments, we determined it to be unsuitable because it contained many non-English example sentences, despite claims in \cite{derczynski2016broad} that it used English language tweets. We replaced Broad Twitter Corpus with another NER dataset, Few-NERD, which contained only English sentences from Wikipedia.

\subsection{Conformal Prediction Method and Hyperparameters}\label{appsub:raps}

To generate conformal prediction sets we used the well-known RAPS procedure \cite{angelopoulos2021raps}, which provably produces sets of lower average size than the \topk method for an equivalent coverage level. RAPS has three hyperparameters: a temperature $T$ used for scaling the model's logits before applying the softmax; a set size regularizer $k_{\mathrm{reg}}$; and a regularization weight $\lambda$. 

Defining $\rho_x(y)=\sum_{y'=1}^M f(x)_{y'}\mathds{1}[f(x)_{y'} > f(x)_y]$ as the total probability mass of the set of labels more likely than $y$ for input $x$, and $o_x(y)=|\{y'\in \mathcal{Y} \mid f(x)_{y'} \geq f(x)_{y} \} |$ as the ranking of $y$ among labels based on softmax scores $f(x)$, the RAPS procedure constructs prediction sets as
\begin{equation}\label{eq:raps}
    \mathcal{C}_{\hat q}(x)  = \{y \mid \rho_x(y) + u\cdot f(x)_y + \lambda(o_x(y) - k_{\mathrm{reg}})^{+} \leq \hat q \}.
\end{equation} 
where $u\sim U[0,1]$ is a uniform random variable. In essence, RAPS defines the score function $s(x, y)$ as a sum of three terms involving the probability mass of more likely classes, the probability of the input class (randomly weighted), and a regularizer that promotes small set sizes by imposing an extra penalty to add classes when $k_{\mathrm{reg}}$ have already been included.

We summarize the hyperparameters of our experiments in Table~\ref{tab:hyper-description}. Also shown is the empirical risk $\hat\alpha$ achieved by the \topk sets, which was then used as $\alpha$ for conformal calibration.

\begin{table}[ht]
    \centering
    \caption{Hyperparameter Settings and Empirical Coverage}
    \label{tab:hyper-description}
    \begin{tabular}{ccccc|c} \toprule
        Dataset & $k$ & $\lambda$ & $T$ & $k_\textrm{reg}$ & $\hat \alpha$ \\ \hline
        ObjectNet  & 3 & 0.5 & 0.002 & 5 & 0.065  \\ \hline
        ObjectNet Ablation  & 3 & 0.3 & 0.005 & 4 & 0.195  \\ \hline
        GoEmotions & 3 & 0.5 & 0.3 & 4 & 0.085 \\ \hline
        GoEmotions Ablation & 3 & 0.5 & 1.3 & 4 & 0.085 \\ \hline
        Few-NERD & 3 & 0.5 & 0.3 & 5 & 0.021 \\ \bottomrule
    \end{tabular}
    \vspace{-10pt}
\end{table}

\section{Human Subject Experiments}\label{app:experiment}

In this section we provide additional details on our human subject experiments to complement the descriptions in Sections \ref{sec:method} and \ref{sec:experiments}. We have released our aggregated experimental data in \href{https://github.com/layer6ai-labs/hitl-conformal-prediction}{this GitHub repository}.

\paragraph{Ethical Considerations} When running any experiment involving human subjects, ethical considerations are of utmost importance. We considered research ethics throughout the experimental process, and followed the ethics guidelines for experiments involving humans published by the NeurIPS 2023 conference organizers, which accords with the standards of ICML 2024. 
In particular, we followed all existing protocols at our institution for such research. Although our institution does not have an internal review board (IRB) process, in its place we took the following steps: understanding the existing process in place at our institution, ensuring that our datasets only contained content appropriate for showing participants, piloting the tests ourselves, informing participants about what data would be collected and what data they would be shown, collecting consent from participants, paying participants a fair wage, and evaluating the demographic distribution of participants in our experiments to be aware of potential biases.

\paragraph{Participant Recruitment and Compensation}  We recruited participants through Prolific, which produces the highest quality data according to scientific comparisons of behavioural data collection platforms \cite{eyal2021dataquality, douglas2023dataquality}. We did not filter the population accepted into our study, other than that we required fluency in English, the language we used for task instructions. When participants voluntarily agreed to join our study, they were randomly assigned to one of three tasks, and one of three treatments. No participant was involved in more than one test. In total, we recruited and paid 600 participants. Compensation followed the guidelines enforced by Prolific. Participants were promised a flat amount of pay based on the median completion time of the test, and were also offered bonus pay proportional to the number of correct answers they gave as a financial incentive to perform well on the tasks. Overall, the average rate of pay was 7.80 GBP/hr, and 1500 GBP was spent on participant compensation in total, exclusive of fees.

All 600 participants consented to the collection and dissemination-in-aggregate of their demographic data. As seen in Table \ref{tab:demographics-performance}, our study population was approximately balanced in terms of gender (58\% Male, 42\% Female), and covered a wide range of ages. Since all the tasks we designed relied only on general human knowledge (and fluency in English), we do not expect any effect of age or gender on task performance (here measured by accuracy). To verify this null hypothesis, we compare the normalized accuracy for each group in Table \ref{tab:demographics-performance}. Because each task and treatment may have a different inherent difficulty, we normalize the accuracy within each task/treatment cohort before aggregating over demographic groups. We show the mean and standard deviation, where a value of 1 indicates the population performed 1 standard deviation above the mean of their cohort. Unsurprisingly, the results are consistent with there being no effect of gender or age on task performance.

\begin{table}[ht]
\centering
\caption{Demographics and Performance of Participants}
\begin{tabular}{clcc}
\toprule
\multicolumn{1}{l}{} & \textbf{Group} & \textbf{\# Participants} & \textbf{Normalized Accuracy} \\ \hline
\multirow{6}{*}{\textbf{Age group}} & $<20$ & 33 & $+0.21\pm1.04$ \\
 & 20-29 & 407 & $+0.06\pm0.93$ \\
 & 30-39 & 95 & $-0.10\pm1.06$ \\
 & 40-49 & 37 & $-0.30\pm1.16$ \\
 & 50-59 & 20 & $-0.35\pm1.15$ \\
 & $\geq 60$ & 8 & $-0.29\pm1.39$ \\ \hline
\multicolumn{1}{l}{\multirow{3}{*}{\textbf{Sex}}} & Male & 345 & $+0.05\pm0.99$ \\
\multicolumn{1}{l}{} & Female & 253 & $-0.07\pm0.99$ \\
\multicolumn{1}{l}{} & Other & 2 & $-0.50\pm1.44$ \\ \bottomrule
\end{tabular}
\label{tab:demographics-performance}
\end{table}

\begin{figure}[h]
\centering
\includegraphics[width=0.5\columnwidth, trim={30, 0, 40, 0}, clip]{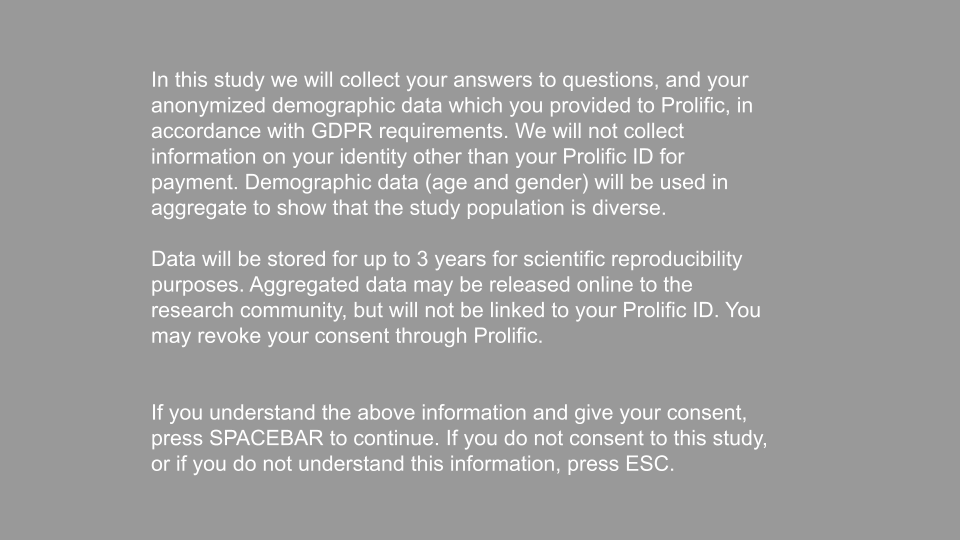}
\caption{Consent screen shown to participants at the start of the experiment.}
\label{fig:consent}
\end{figure}

\begin{figure}[h]
\centering
\includegraphics[width=0.5\columnwidth, trim={30, 0, 40, 0}, clip]{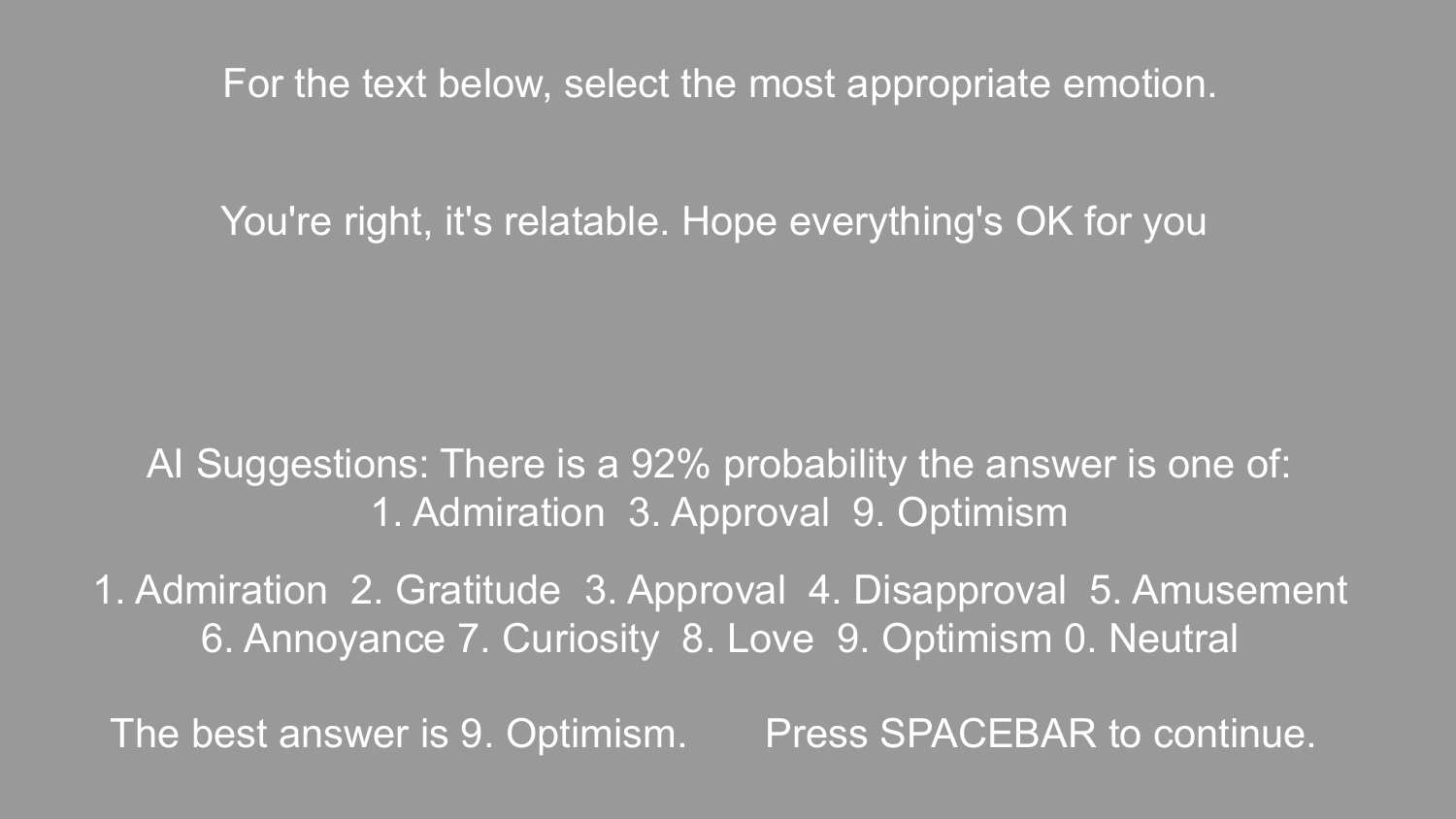}
\caption{Main trial screen shown to participants for GoEmotions with conformal set treatment. The correct answer is given only after the participant responds.}
\label{fig:goemotions-trial}
\end{figure}

\begin{figure}[t]
\centering
\includegraphics[width=0.49\columnwidth]{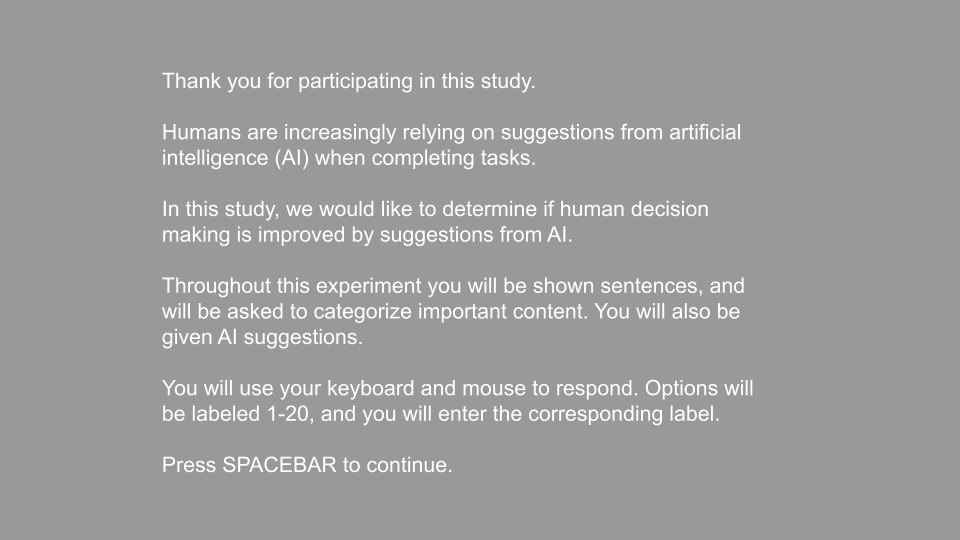}
\includegraphics[width=0.49\columnwidth]{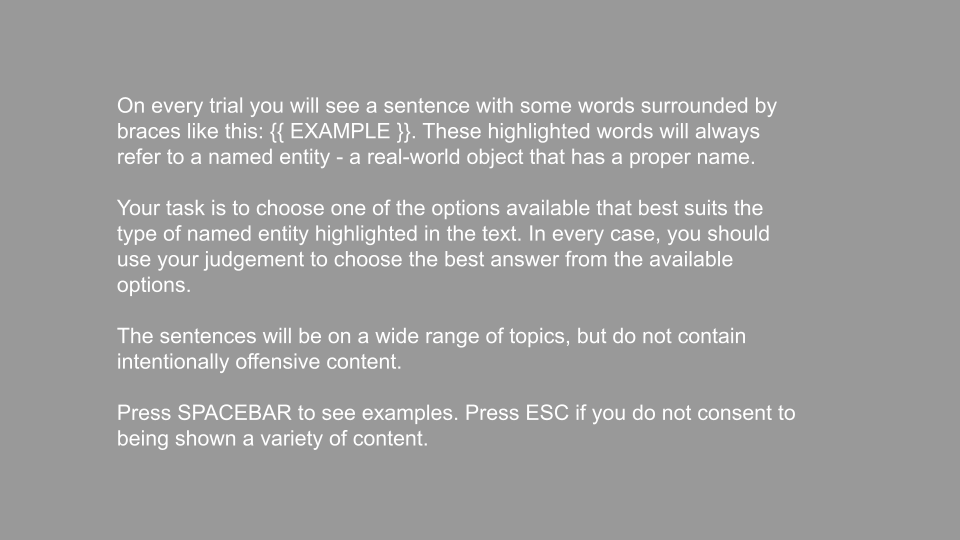}
\includegraphics[width=0.49\columnwidth]{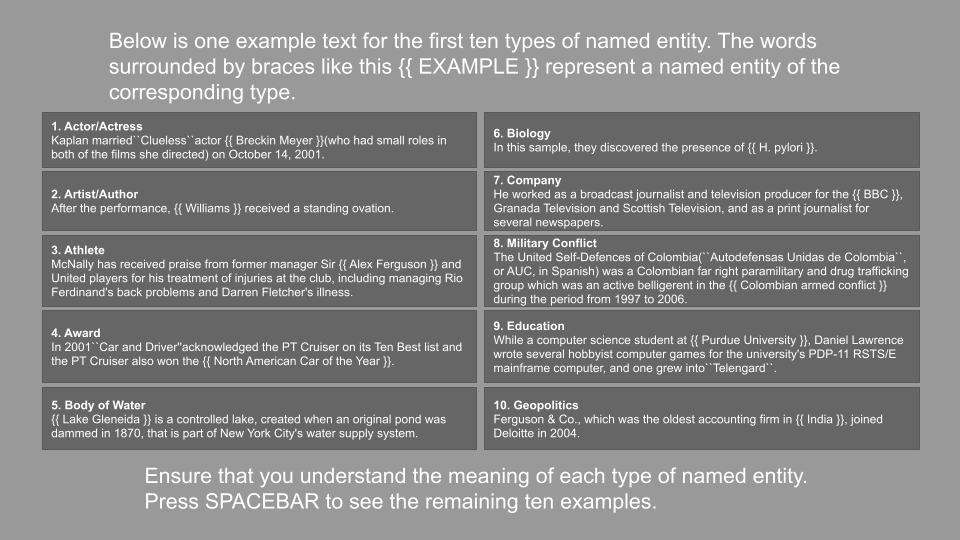}
\includegraphics[width=0.49\columnwidth]{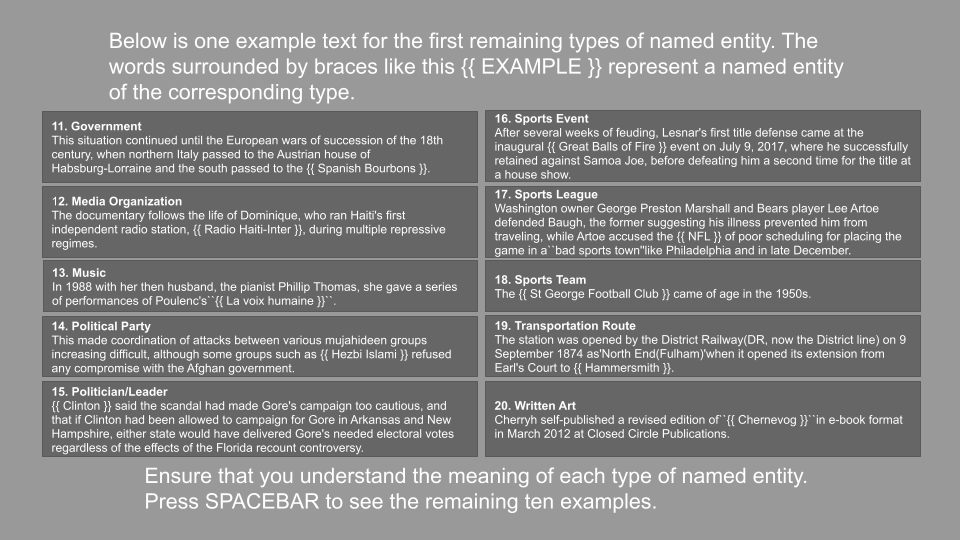}
\includegraphics[width=0.49\columnwidth]{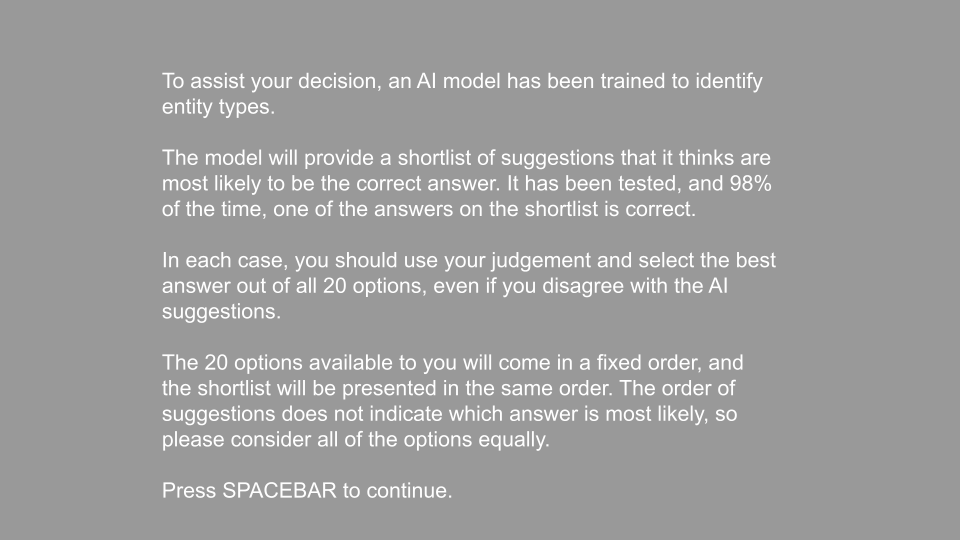}
\includegraphics[width=0.49\columnwidth]{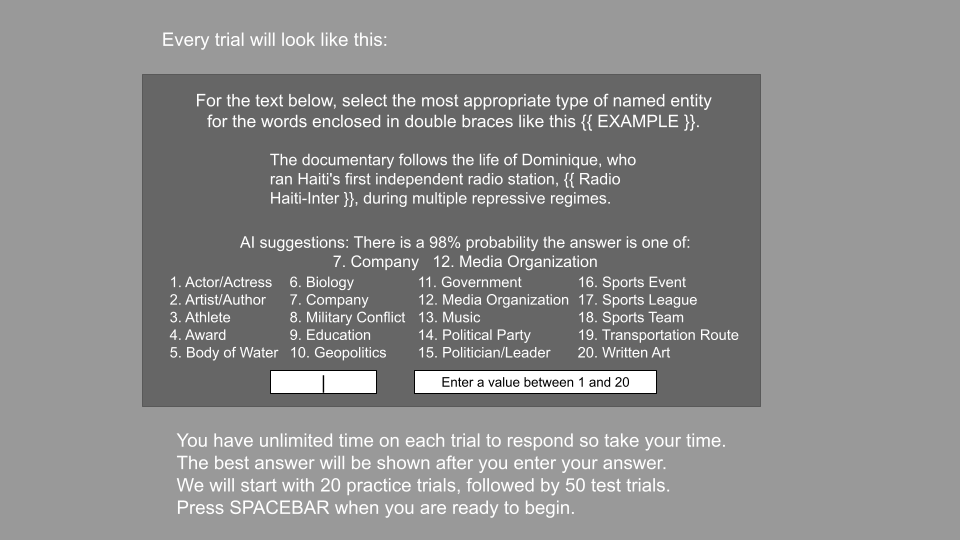}
\includegraphics[width=0.49\columnwidth]{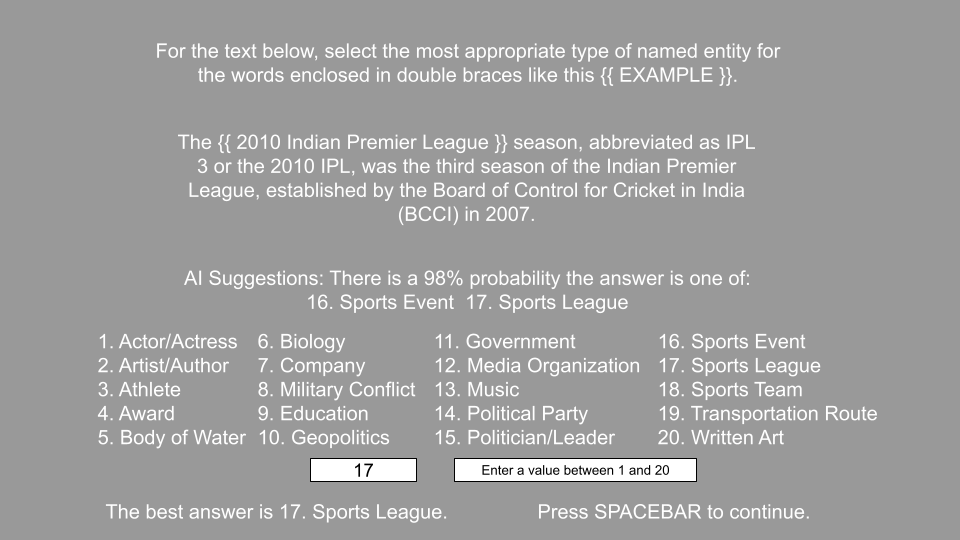}
\includegraphics[width=0.49\columnwidth]{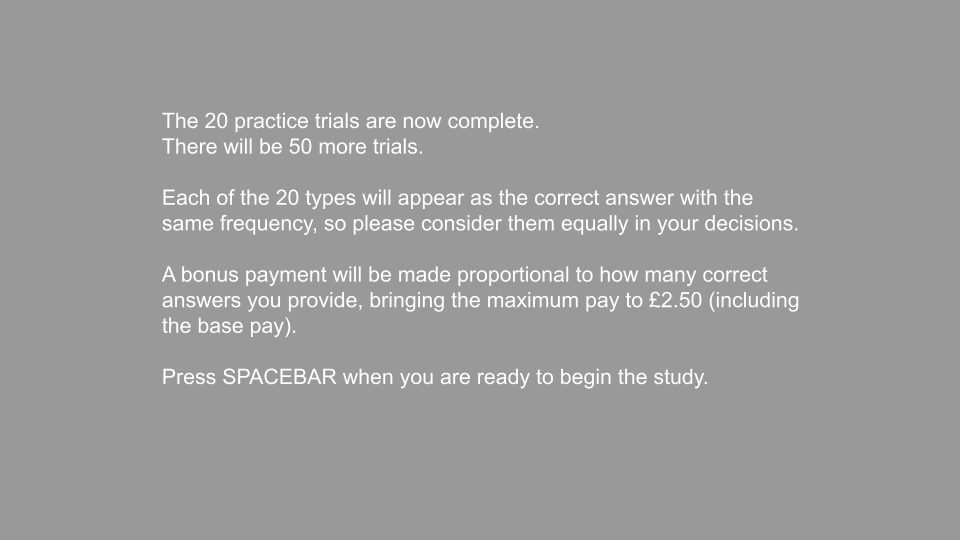}
\caption{Screens displayed to participants during our experiment using conformal sets on Few-NERD. Other experiments followed the same template. The bottom left shows the format of the practice and test trials, with the correct answer text shown only after a participant entered their response. The slide in the first column, third row was not shown to the control group, and they were not provided any prediction set information in the practice and test trials.}
\label{fig:experiment_screens}
\end{figure}

\paragraph{Experiment Details}
Participants were required to use a desktop or laptop computer to complete our experiment (not a mobile device or tablet). To begin, participants were shown a statement on the data we were collecting, including information on how we planned to store, disseminate, and release that data publicly for scientific purposes. Participants had the option to consent to this treatment, or remove themselves from the study (see \autoref{fig:consent}). We did not collect Personal Identifiable Information (PII) such as name, address, birth date, governmental identification numbers, or banking information. There were no potential risks of participating that we needed to disclose to participants. Then, instructions were shown introducing the task, and we provided labelled examples of each class. Next, participants went through 20 practice trials of the same format as the actual test so they could further learn about the dataset and classes. Example and practice stimuli were taken from $\mathcal{D}_{\mathrm{test}}$, were the same for all participants, and were never reselected as a test trial. Finally, the test trials were conducted using randomized samples from $\mathcal{D}_{\mathrm{test}}$; for both GoEmotions and Few-NERD $m=50$ trials were used, while for ObjectNet we assigned $m=100$ trials because they could be completed more quickly. Our experimental data is based only on the test trials. \autoref{fig:experiment_screens} shows the screens that were displayed to participants in the Few-NERD experiment; the other two experiments followed the same template, and examples of their main trial screens are given in \autoref{fig:objectnet-trial} and \autoref{fig:goemotions-trial}.

Each trial showed the participant a stimulus $x$ and all possible classes $\mathcal{Y}$, forcing the participant to choose one class with unlimited time to make their choice. For ObjectNet, the stimuli were only shown for 0.22 s to increase the difficulty of the task. Participants were given the correct answer after they responded, both during the practice and testing phase. Training of this kind has been shown to have a large positive effect on the quality of data collected \cite{mitra2015comparing}. It is possible that the participant may learn about the task while taking the test and change the way they answer to maximize accuracy. We see this as a desirable effect; ultimately we want participants to perform as well as possible. Additional justifications for providing correct answers during the testing phase are given in \cite{stein2023exposing}.

Due to inherent differences in the datasets, there were small adjustments made to each task. In all cases $\mathcal{D}_{\mathrm{cal}}$ and $\mathcal{D}_{\mathrm{test}}$ were exactly class balanced through stratification. For GoEmotions, $M=10$ classes were used, while $M=20$ was chosen for the other tests. Because we implement a forced choice test, increasing $M$ increases the difficulty of the task, as additional possible answers can confound the true label. In preliminary testing we found GoEmotions to be the hardest task already at $M=10$, and this is borne out in \autoref{fig:main-acc}. For the tests with $M=20$, randomly sampling stimuli from $\mathcal{D}_{\mathrm{test}}$ can result in a very skewed distribution of classes which subsequently adds variance to the difficulty of the test seen by any individual (for example, \autoref{fig:obj-acc-by-class} shows that there is variation in the accuracy of responses by class for ObjectNet). Hence, for these tests we performed stratified sampling from $\mathcal{D}_{\mathrm{test}}$ for the stimuli shown to each participant. Our GoEmotions experiment was conducted in a manner that could be completed very quickly if the participant did not attempt to provide accurate responses. To prevent delinquency, we incorporated two attention checks in accordance with the design principles laid out by Prolific. Participants were rejected from the study and replaced if they failed both attention checks, as planned in our pre-registration. In total, one participant failed both attention checks.
\clearpage
\section{Additional Analysis}\label{app:analysis}

In this Appendix, we provide additional analysis on the data we collected, in support of Section \ref{sec:results}.

\begin{figure}[t]
\centering
\includegraphics[width=0.7\columnwidth, trim={0, 0, 0, 0}, clip]{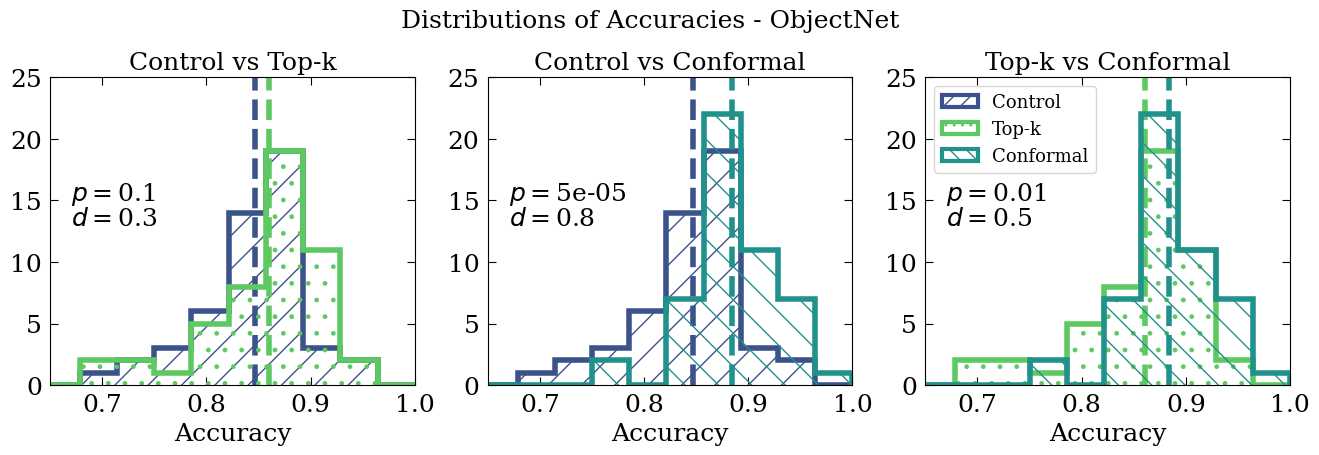}
\includegraphics[width=0.7\columnwidth, trim={0, 0, 0, 0}, clip]{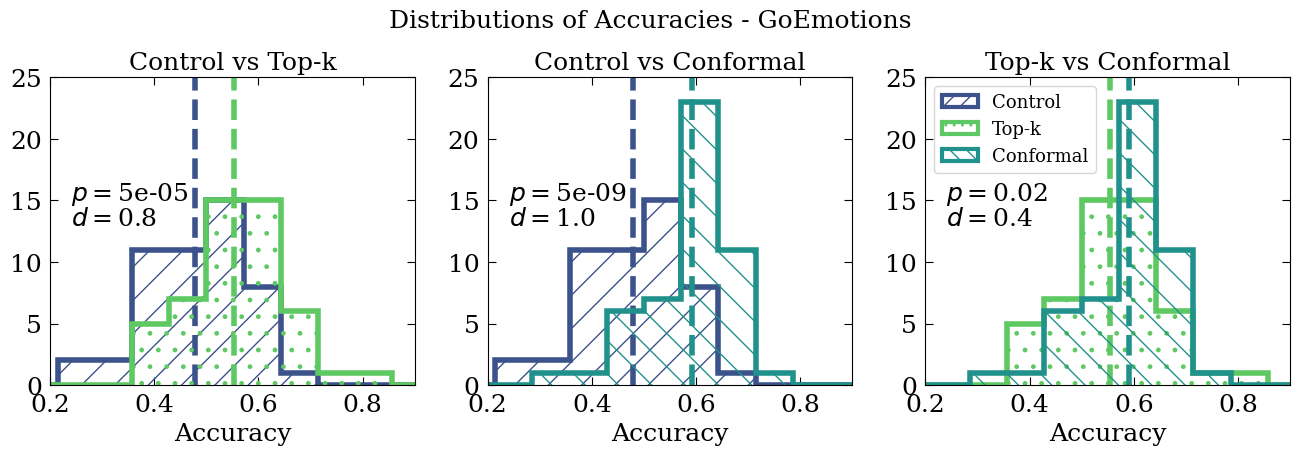}
\includegraphics[width=0.7\columnwidth, trim={0, 0, 0, 0}, clip]{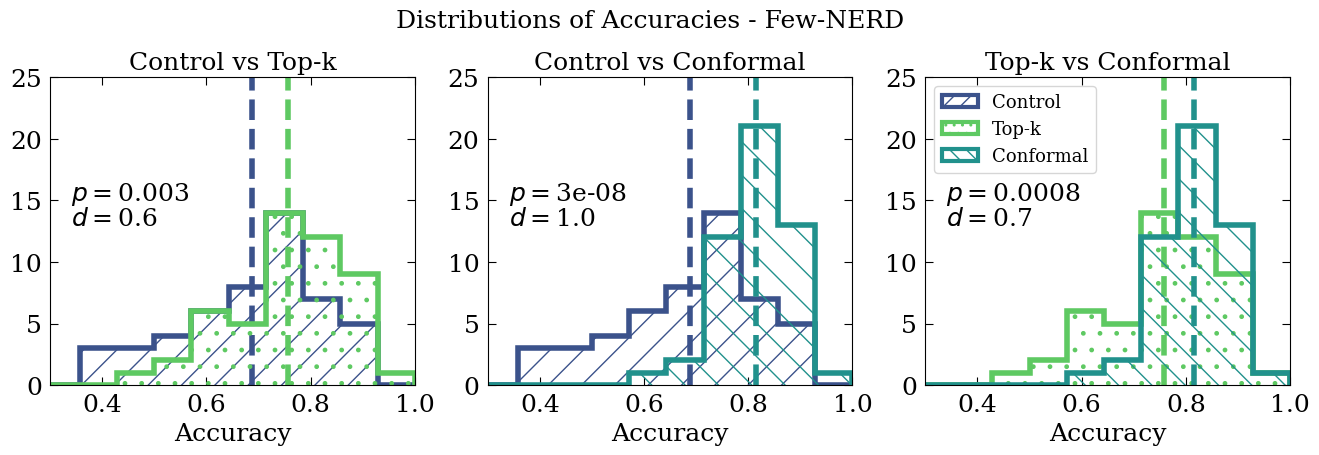}
\caption{Histograms of accuracy observations across three treatments for each task. Vertical lines indicate the mean of each distribution, $p$ indicates the $p$-value given by Welch's $t$-test, and $d$ is the effect size given by Cohen's $d$.}
\label{fig:histogram-acc}
\end{figure}

\autoref{fig:main-acc} summarized our main experimental results regarding human accuracy on our tasks. In \autoref{fig:histogram-acc} we show the histograms of $N=50$ accuracy observations for each task, making pairwise comparisons between treatments. According to the significance tests compiled in \autoref{tab:p-values-acc}, the conformal treatment shows statistically significant improvement in accuracy compared to the other treatments on each of the tasks.

\begin{figure}[t]
\centering
\includegraphics[width=0.7\columnwidth, trim={0, 0, 0, 0}, clip]{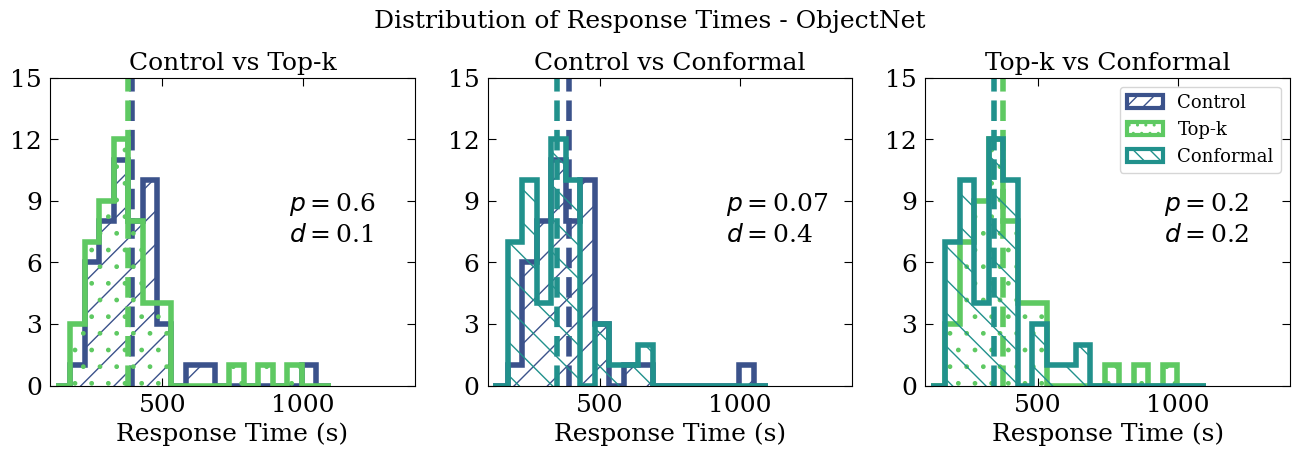}
\includegraphics[width=0.7\columnwidth, trim={0, 0, 0, 0}, clip]{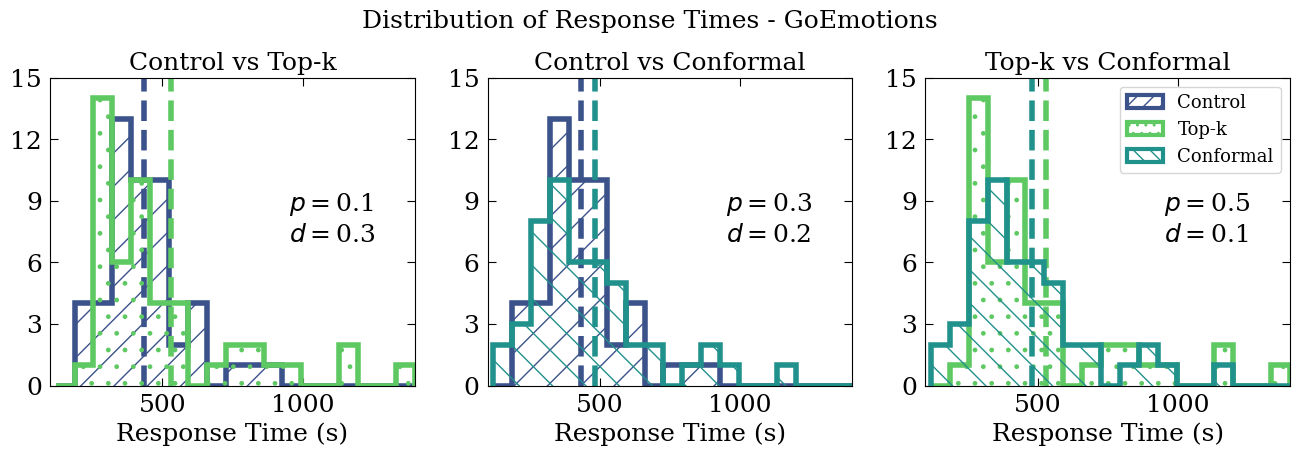}
\includegraphics[width=0.7\columnwidth, trim={0, 0, 0, 0}, clip]{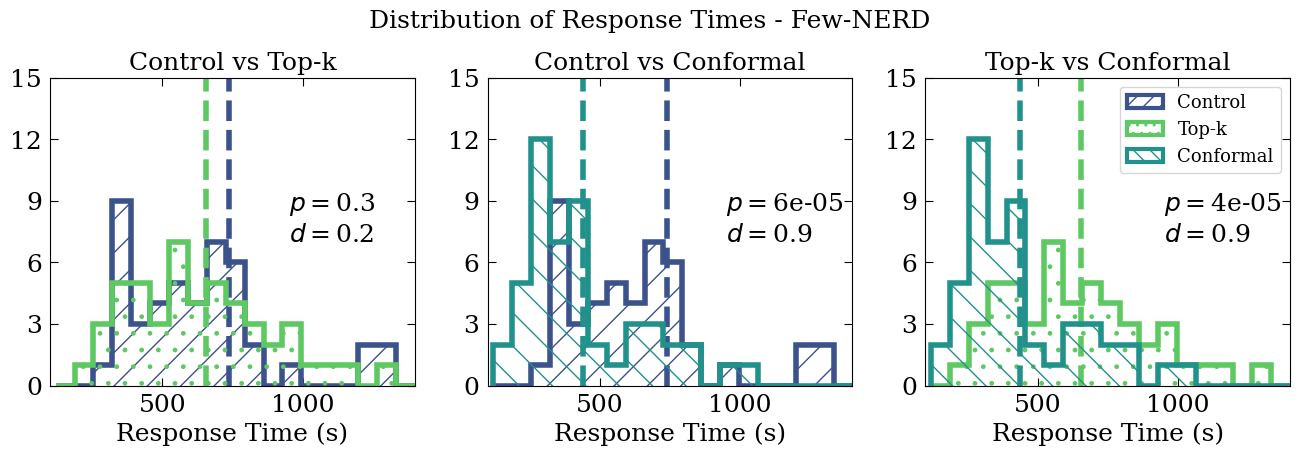}
\caption{Histograms of response time observations across three treatments for each task. Vertical lines indicate the mean of each distribution, $p$ indicates the $p$-value given by Welch's $t$-test, and $d$ is the effect size given by Cohen's $d$.}
\label{fig:histogram-time}
\vspace{-10pt}
\end{figure}

Our results on response times were displayed in \autoref{fig:main-time}. Once again we show the histograms of $N=50$ average response time observations for each task in \autoref{fig:histogram-time}. Unlike for accuracy, we do not see strong trends for prediction sets increasing or decreasing decision speed across all cases, and in most cases we do not see significant differences between the means of the distributions (\autoref{tab:p-values-time}).

In our pre-registration we stated that data analysis would be conducted using the one-way ANOVA method. In the ensuing time since pre-registration, and based on the discussion in Section \ref{sec:method} we found that applying Welch's $t$-test was a suitable, and simpler statistical approach, and therefore used it for our final analysis instead. 

\newpage

\begin{figure*}[h!]
    \centering
    \includegraphics[width=0.49\linewidth]{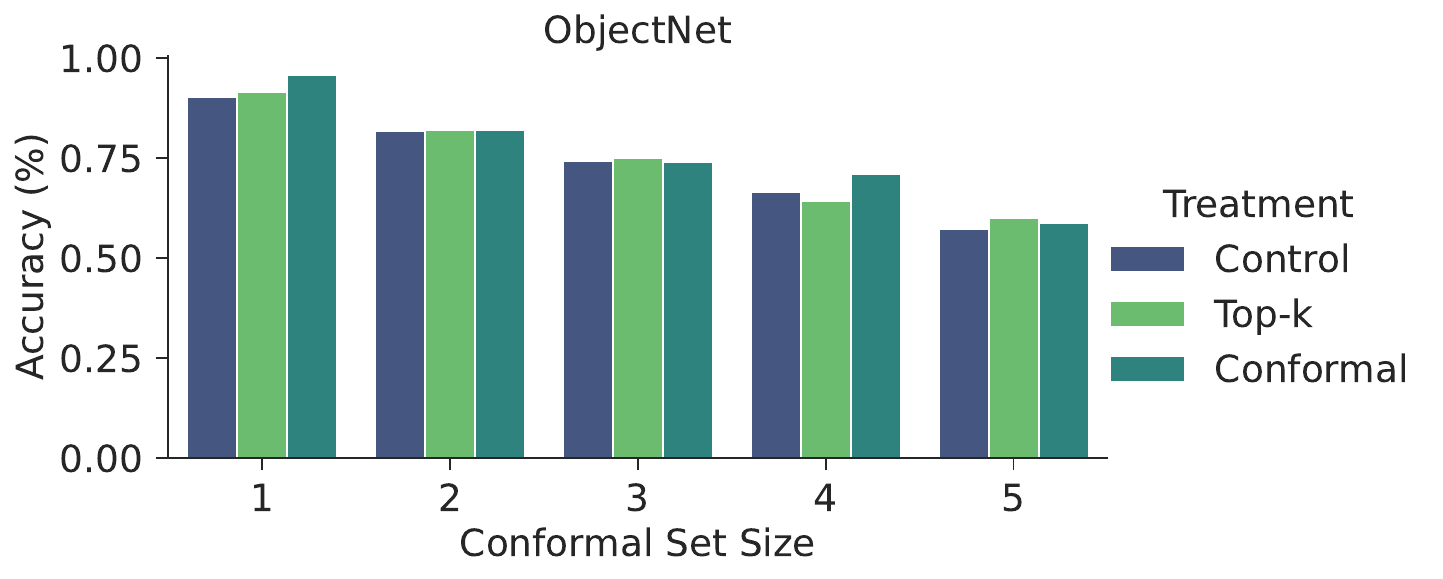}
    \includegraphics[width=0.49\linewidth]{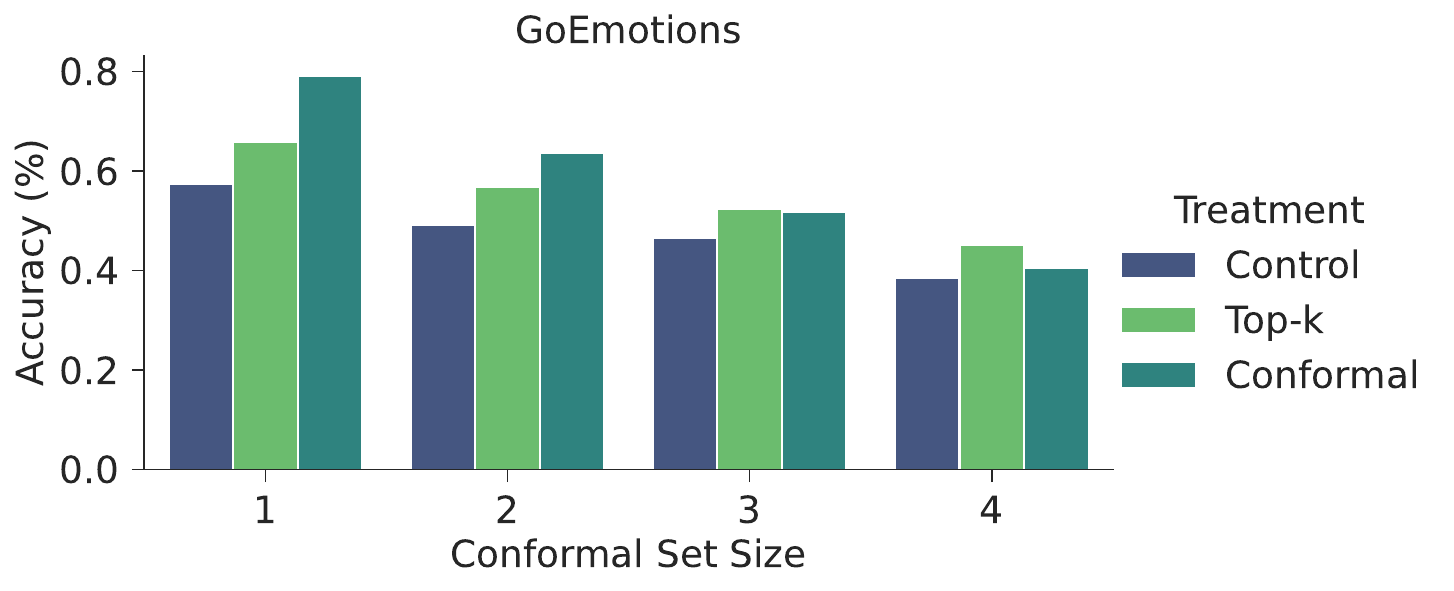}
    \caption{Human accuracy by difficulty of examples (conformal set size).}
    \label{fig:acc-conf}
\end{figure*}

To complement \autoref{fig:acc-set-size} in the main text for Few-NERD, in \autoref{fig:acc-conf} we present the human accuracies conditioned on conformal set size for both ObjectNet and GoEmotions. Similar to the observations made for Few-NERD, the provision of prediction sets tended to enhance human accuracy across uncertainty levels, but demonstrated the largest impact on accuracy for singleton sets.

\begin{figure*}
    \centering
    \includegraphics[width=0.45\linewidth, trim={0, 0, 0, 0}, clip]{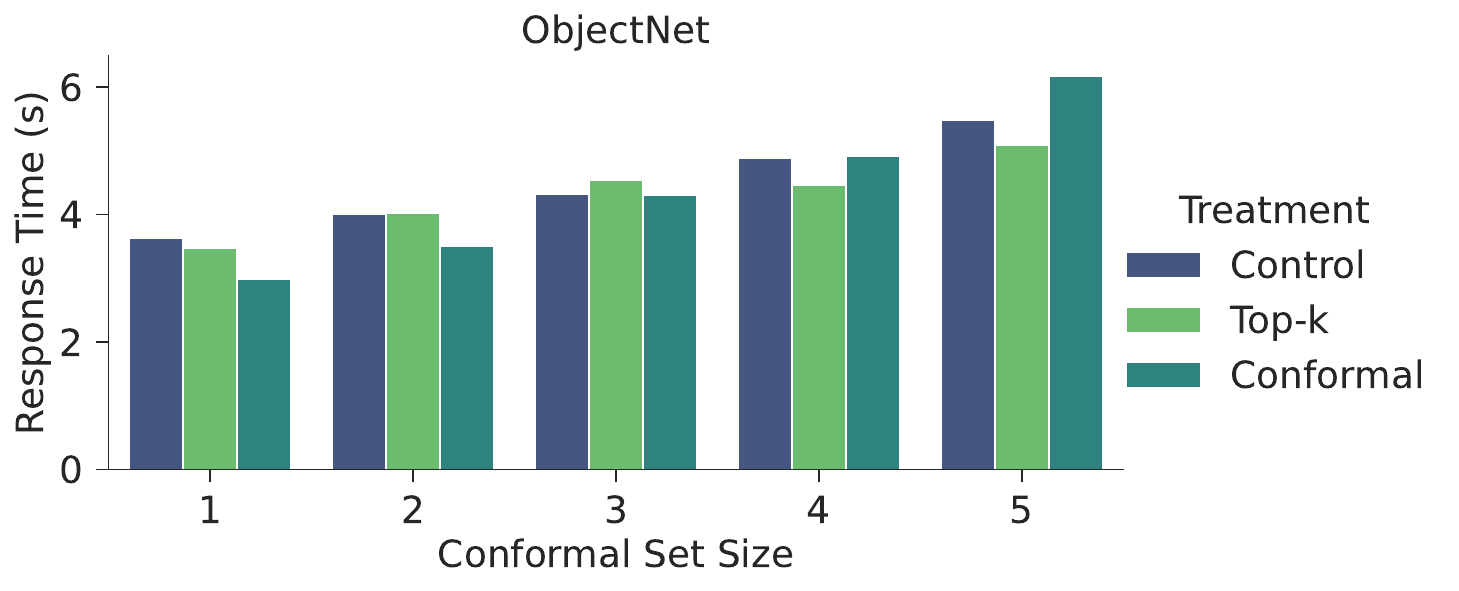}
    \includegraphics[width=0.45\linewidth, trim={0, 0, 0, 0}, clip]{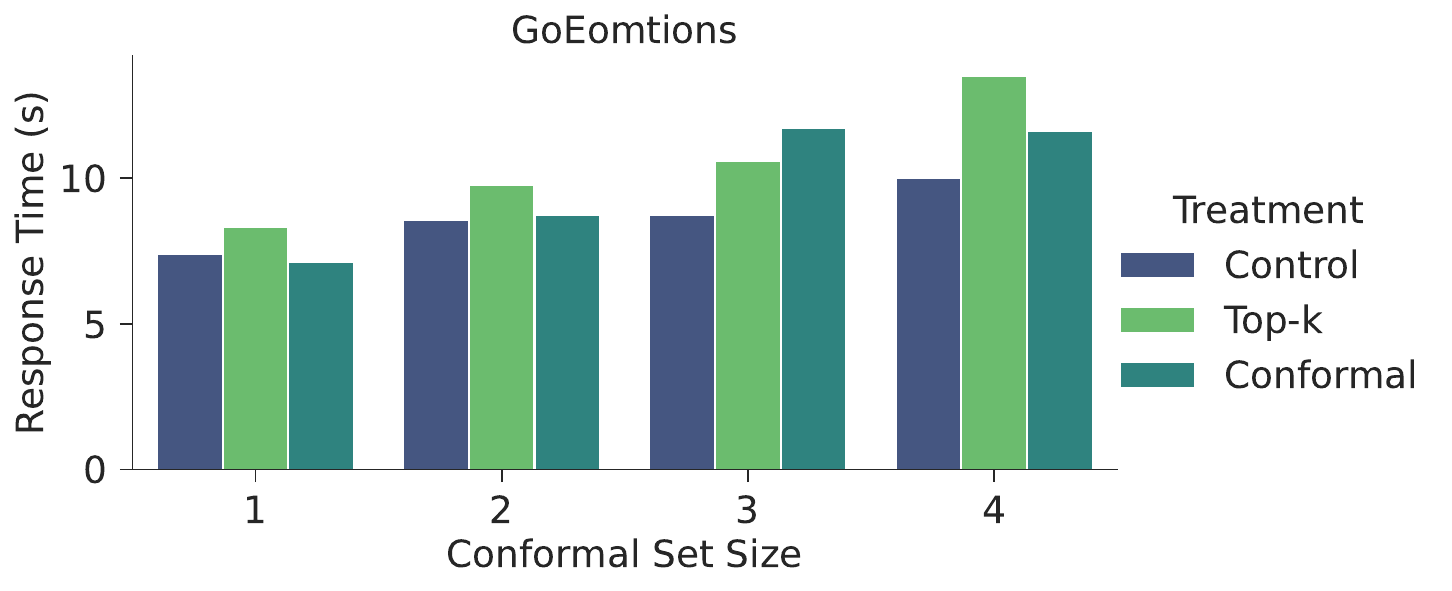}
    \caption{Human response time by difficulty of examples (conformal set size).}
    \label{fig:time-conf}
\end{figure*}

Similarly, \autoref{fig:time-conf} completes what was shown in \autoref{fig:time-set-size} for the remaining datasets. The response time trend observed for GoEmotions and ObjectNet aligns with our findings for Few-NERD: longer response times are consistently observed across all treatment groups when dealing with more uncertain samples, as judged by the models, while singleton conformal sets improve prediction efficiency the most. 

\begin{figure*}
    \centering
    \includegraphics[width=0.4\linewidth, trim={0, 0, 0, 0}, clip]{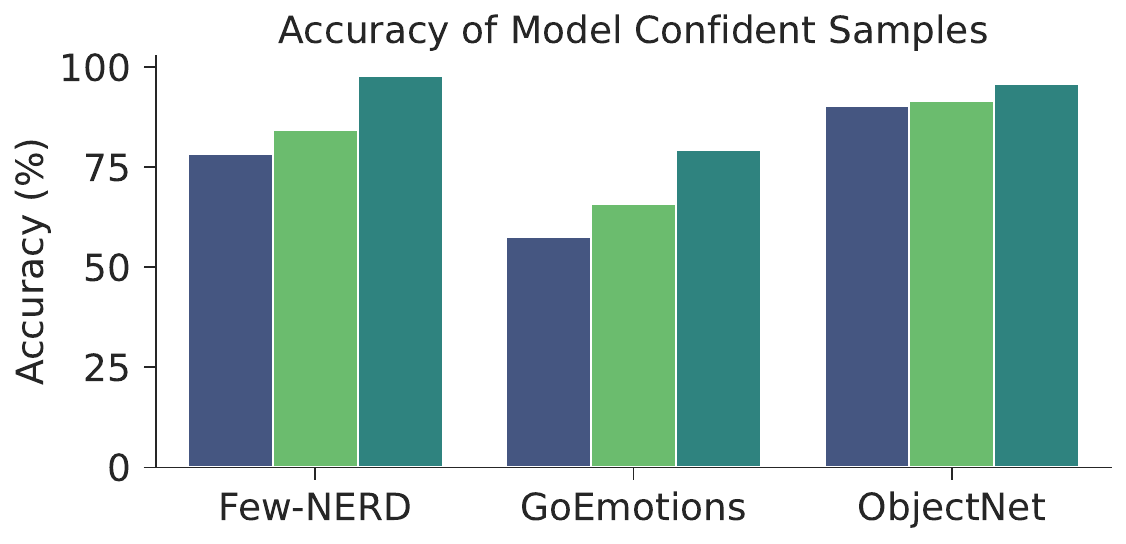}
    \includegraphics[width=0.5\linewidth, trim={0, 0, 0, 0}, clip]{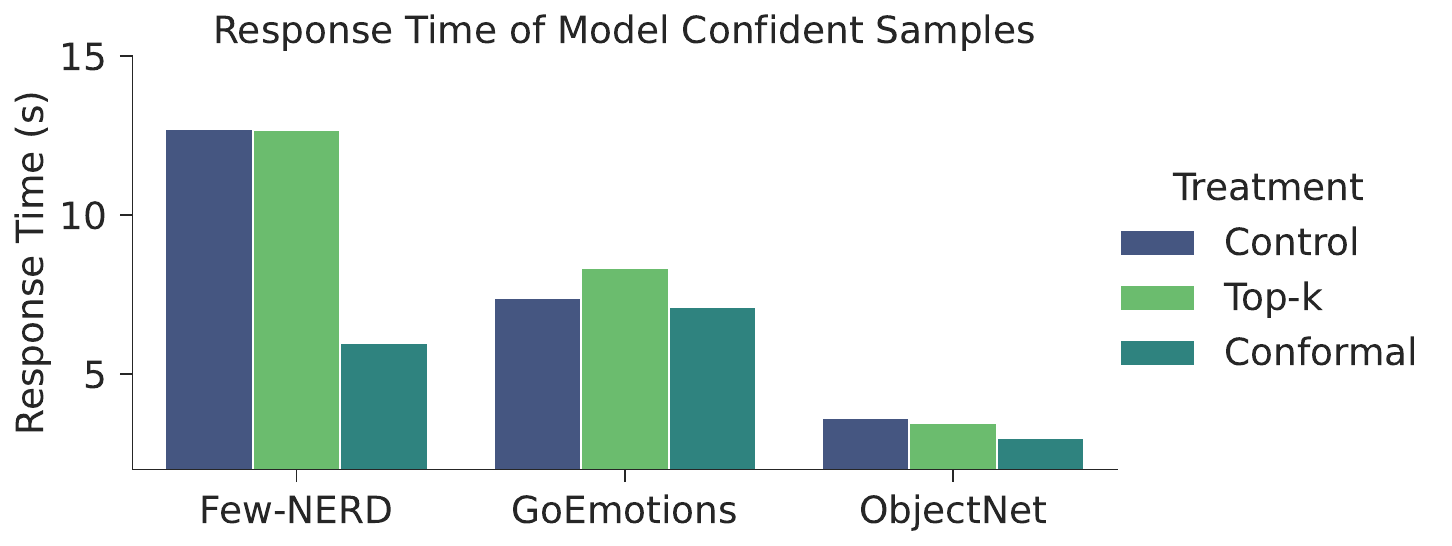}
    \caption{Human accuracy and response time on samples with low uncertainty according to the models (i.e. singleton conformal sets).}
    \label{fig:singleton-sets}
\end{figure*}

The positive effect of singleton conformal sets is further emphasized in \autoref{fig:singleton-sets}.

\end{document}